%% file: main.tex
\documentclass[journal]{IEEEtran}
\input{preamble}
\begin{document}
\title{Ziv Meets Sumcheck: Zero-Knowledge Proofs\\ of Floating-Point LLM Inference}
\author{Hassan~Khodaiemehr,~\IEEEmembership{Member,~IEEE,}
Khadijeh~Bagheri,
and~Chen~Feng,~\IEEEmembership{Member,~IEEE}%
\thanks{H. Khodaiemehr, K. Bagheri, and C. Feng are with the School of Engineering, University of
British Columbia (Okanagan Campus), Kelowna, BC, Canada.
E-mail: \{hassan.khodaiemehr, khadijeh.bagheri, chen.feng\}@ubc.ca.}}
\markboth{Preprint}%
{Khodaiemehr \MakeLowercase{\textit{et al.}}: Ziv Meets Sumcheck}
\maketitle

\begin{abstract}
Zero-knowledge proofs can show that a neural network produced a given output without revealing the
network, but current systems for language models prove integer approximations of the model rather
than the floating-point arithmetic that is actually deployed. Proving floating-point matrix
multiplication has been regarded as hard, because rounding after every addition makes the result
depend on the order of the additions. We avoid this obstacle by proving the correctly rounded
product, in which every dot product is computed exactly and rounded once. This result is
independent of the hardware and at least as accurate as standard kernels. We give an interactive
proof for this relation whose cost is close to that of integer matrix multiplication, with roughly a
thousand times fewer committed values than per-operation floating-point circuits need constraints, extend it to every
layer of a transformer, and give a privacy construction for the complete argument whose zero knowledge we prove under explicit hypotheses. We implement the system and
prove the full forward pass of GPT-2 small, including all twelve blocks and the output over the whole
vocabulary, in 19 minutes on two processor cores; verification takes 21 seconds. The verifier
rejects all 236 attacks we tested, and the model's outputs agree with high-precision inference at
least as closely as standard 16-bit inference does.
\end{abstract}

\begin{IEEEkeywords}
Zero-knowledge proofs, verifiable machine learning, floating-point arithmetic, correct rounding,
sumcheck protocol, large language models.
\end{IEEEkeywords}

\section{Introduction}\label{sec:intro}
\IEEEPARstart{S}{uccinct} arguments for machine-learning inference have progressed from convolutional
networks~\cite{zkcnn21} to large language models~\cite{zkllm24,zkgpt25,deepprove26,openllm26,joltatlas26,nanozk26},
up to mixture-of-experts models with about $4\cdot10^{11}$ parameters~\cite{flashprove26,sash26}. All of
these systems arithmetize an \emph{integer} model, whereas deployed models run in \bfs{} or 8-bit
floating point with binary32 accumulation. The authors of~\cite{flashprove26} note that floating-point
outputs may diverge from the proved model, \cite{sash26} lists ``matching served arithmetic'' as open
and reports a 7\% rise in perplexity under the proved arithmetic, and for training Peign\'e-Lefebvre
et al.~\cite{peigne26} pose ``algebraic reductions for floating-point GEMM verification'' as open
problem OP-1, because floating-point addition is not associative and exact Freivalds checks and
sumcheck therefore fail.

Known proofs of IEEE-style arithmetic build a gadget for each floating-point
operation~\cite{garg22,ernstberger24,mystique21}, about $64$ constraints per single-precision
multiplication~\cite{ernstberger24}, hence $\Theta(mnk)$ gadgets for an $m\times k$ by $k\times n$
product; ZIP~\cite{zip25} reports $34.5$ hours for the linear layers of a model with $1.1\cdot10^7$
parameters. Integer zkML avoids the cubic term with Thaler's sumcheck~\cite{thaler13}, which needs the
product to be a polynomial identity, and a kernel that rounds after every addition computes a
function of the \emph{order} of the additions.

\begin{figure}[t]
\centering
\begin{tikzpicture}[font=\scriptsize, node distance=3mm and 3mm,
  box/.style={draw, rounded corners=2pt, align=center, inner sep=2pt, minimum height=8mm, text width=2.35cm},
  arr/.style={-{Stealth[length=1.6mm]}, thick}]
\node[box] (ops) {operands $A,B$:\\ limbs $X$, buckets $g$};
\node[box, right=of ops] (poly) {$A(z)B(z)=\sum_t D^{(t)}z^t$\\ (\cref{sec:buckets})};
\node[box, right=of poly] (sc) {random $z$, one\\ matmul sumcheck};
\node[box, below=of sc] (chain) {carry chain: exact\\ $N=\sum_t D^{(t)}\beta^t$};
\node[box, left=of chain] (gad) {rounding gadget\\ (\cref{sec:gadget})};
\node[box, left=of gad] (out) {output\\ $C=\RN(AB+b)$};
\draw[arr] (ops) -- (poly); \draw[arr] (poly) -- (sc); \draw[arr] (sc) -- (chain);
\draw[arr] (chain) -- (gad); \draw[arr] (gad) -- (out);
\end{tikzpicture}
\caption{Structure of the protocol: the cubic part is one sumcheck over the inner index; everything
specific to floating point happens once per output entry.}
\label{fig:overview}
\end{figure}
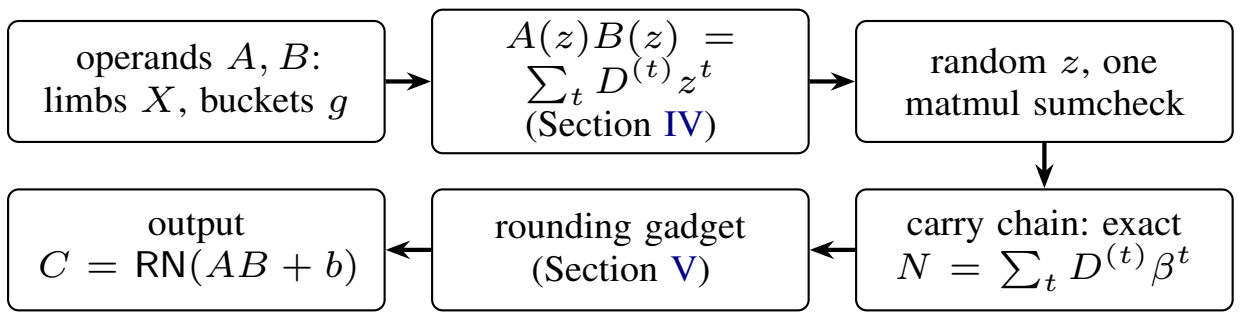

We change which floating-point function is proved. For floating-point $A$, $B$ and a bias $b$ let
\[
  \CR(A,B,b)_{ij} \;=\; \RN\Bigl(\,\textstyle\sum_{l} A_{il}B_{lj} + b_j\Bigr),
\]
the exact sum rounded once to nearest, ties to even: the requirement that IEEE~754 places on every
basic operation~\cite{ieee754}, applied to the whole dot product~\cite{muller18,ogita05,demmel13}.
It is \emph{canonical}, independent of accumulation order, blocking, hardware and batch size, whose
effect on native kernels breaks bit-reproducibility~\cite{cankaya26}; it is a \emph{legitimate
execution}, with error at most half a unit in the last place and agreement with native kernels on all
but about $10^{-4}$ of the entries; and it is \emph{algebraic}, an integer identity followed by one
rounding per output, $mn$ roundings instead of $mnk$. As in Ziv's strategy for correctly rounded
functions~\cite{ziv91}, we compute with a certified bound, test whether the rounding is determined,
and fall back to exact computation only otherwise, both to obtain the canonical value from a native
kernel and as the design principle of the proof (\cref{fig:overview}).

Three problems are solved: exact products are integers of hundreds of bits, which we split into short
limbs by \emph{exponent buckets} so that one sumcheck at a random $z$ checks the product
(\cref{sec:buckets}); field equations must hold over the integers, which limb bounds and a
range-checked carry chain ensure (\cref{lem:gadget}); and the rounding must be exact, including ties,
binade boundaries and underflow, which one residual characterizes (\cref{lem:round,sec:gadget}).
\Cref{tab:contrib} lists the contributions. All numbers are measured, none is projected: native
kernels differ from the canonical result at $9.5\cdot10^{-5}$ of $8.2\cdot10^7$ GPT-2 product outputs, as
often as two native orders differ from each other, with indistinguishable perplexity; the protocol
commits $9.4\cdot10^6$ cells per token against $8.1\cdot10^9$ constraints for per-operation gadgets; all 343
products of GPT-2 are proved in 372~s on two cores; and the complete model is proved with the privacy
construction in 18.7~minutes, verified in 21~s, with every one of 236 attacks rejected.

\begin{table}[t]
\centering\footnotesize\setlength{\tabcolsep}{2pt}
\caption{Contributions.}\label{tab:contrib}
\begin{tabular}{@{}p{1.45cm}p{5.5cm}l@{}}
\toprule
contribution & content & where\\
\midrule
relations & CR matrix product with fused bias; order-agnostic error bound and rounding test for native
kernels; Ziv-style computation of the canonical result & \cref{sec:kernels}\\
protocol & polynomial IOP for $C=\CR(A,B,b)$: exponent bucketing, integer lifting, exact rounding
gadget; complete and sound under explicit field-size conditions & \cref{thm:main}\\
block & every operation of a transformer block (canonical LayerNorm and softmax, lookups, maximum by
selection), chained through views & \cref{thm:block}\\
privacy, full model & hiding commitment, masked sumchecks, zero sums, hidden LogUp sums, embedding and
output checks; ZK under explicit, unproved hypotheses & \cref{thm:zk}\\
evaluation & reference and native implementations on GPT-2 with real weights & \cref{sec:eval}\\
\bottomrule
\end{tabular}
\end{table}

\emph{Scope and limitations.} (i)~The provers are unaudited prototypes with $O(\sqrt N)$ proofs. Zero
knowledge of the complete transcript is proved only under a masking hypothesis and a leaf-entropy
condition (\cref{def:hyp,thm:zk}); rejecting attacks is no evidence of privacy. The $2^{-85}$ soundness
bound is interactive; we give no bound for the Fiat--Shamir implementation or in the quantum
random-oracle model. The relation is that of the padded (widened) architecture with published,
uncertified unary tables, and parameters are committed afresh in every proof. (ii)~The canonical model
rounds after every operation, $1.2$ times less accurate than reference \bfs{}; formats are public and
were chosen from the input. (iii)~Comparisons with gadgets count cells against constraints. (iv)~Serving
the canonical semantics needs a wider accumulator or a repair pass. Outlined proofs and secondary
experiments are in the supplementary material.

\input{theory}

\input{eval}

\bibliographystyle{IEEEtran}
\bibliography{refs}

\clearpage
\setcounter{section}{0}\setcounter{subsection}{0}\setcounter{table}{0}\setcounter{figure}{0}
\setcounter{equation}{0}\setcounter{lemma}{0}\setcounter{theorem}{0}\setcounter{remark}{0}
\setcounter{proposition}{0}\setcounter{corollary}{0}\setcounter{definition}{0}
\renewcommand{\thesection}{S-\Roman{section}}\renewcommand{\theHsection}{S\arabic{section}}
\renewcommand{\thetable}{S\arabic{table}}\renewcommand{\theHtable}{S\arabic{table}}
\renewcommand{\thefigure}{S\arabic{figure}}\renewcommand{\theHfigure}{S\arabic{figure}}
\renewcommand{\theremark}{S\arabic{remark}}\renewcommand{\theHremark}{S\arabic{remark}}
\renewcommand{\thelemma}{S\arabic{lemma}}\renewcommand{\theHlemma}{S\arabic{lemma}}
\renewcommand{\theequation}{S\arabic{equation}}\renewcommand{\theHequation}{S\arabic{equation}}
\renewcommand{\theHtheorem}{S\arabic{theorem}}\renewcommand{\theHproposition}{S\arabic{proposition}}
\renewcommand{\theHcorollary}{S\arabic{corollary}}\renewcommand{\theHdefinition}{S\arabic{definition}}
\twocolumn[{\begin{center}\vspace*{2pt}
{\LARGE Supplementary Material}\\[6pt]
{\large Ziv Meets Sumcheck: Zero-Knowledge Proofs of Floating-Point LLM Inference}\\[4pt]
{\normalsize Hassan Khodaiemehr, Khadijeh Bagheri, and Chen Feng}\\[10pt]
\begin{minipage}{0.9\textwidth}\small\itshape References to sections, lemmas, theorems, tables, equations
and bibliography entries without the prefix S refer to the main paper.\end{minipage}\vspace{14pt}
\end{center}}]
\section{Proofs}\label{sup:proofs}
\input{supp_proofs}
\section{Zero Knowledge: Further Details}\label{sup:zk}
\input{supp_zk}
\input{supp_ext}

\section{Additional Experiments}\label{sup:eval}
\input{supp_eval}
\input{appendix}
\end{document}

%% file: preamble.tex
\usepackage{amsmath,amssymb,mathtools}

\usepackage{amsthm}
\usepackage{booktabs,graphicx,xcolor,array,tabularx}
\usepackage[shortlabels]{enumitem}
\setlist{leftmargin=*,itemsep=0pt,topsep=2pt,parsep=0pt}
\usepackage{cite}
\usepackage{tikz}
\usetikzlibrary{arrows.meta,positioning}
\usepackage[colorlinks=true,linkcolor=blue!50!black,citecolor=blue!50!black,urlcolor=blue!50!black]{hyperref}
\usepackage[capitalise,noabbrev]{cleveref}
\usepackage{microtype}
\allowdisplaybreaks

\AtBeginDocument{%
 \setlength{\abovedisplayskip}{4pt plus 2pt minus 1pt}\setlength{\belowdisplayskip}{4pt plus 2pt minus 1pt}%
 \setlength{\abovedisplayshortskip}{2pt plus 1pt}\setlength{\belowdisplayshortskip}{3pt plus 1pt minus 1pt}}

\newtheorem{theorem}{Theorem}
\newtheorem{lemma}{Lemma}
\newtheorem{proposition}{Proposition}
\newtheorem{corollary}{Corollary}

\theoremstyle{definition}
\newtheorem{definition}{Definition}
\newtheorem{remark}{Remark}
\newcommand{\Fl}{\mathsf{Fl}}
\newcommand{\Fp}{\mathbb{F}_p}
\newcommand{\K}{\mathbb{K}}
\newcommand{\Z}{\mathbb{Z}}
\newcommand{\R}{\mathbb{R}}
\newcommand{\RN}{\mathsf{RN}}
\newcommand{\eq}{\mathsf{eq}}
\newcommand{\CR}{\mathrm{CR}}
\newcommand{\Xmax}{X_{\max}}
\newcommand{\Emax}{E_{\max}}
\newcommand{\bfs}{\texttt{bf16}}
\newcommand{\tenc}{\mathcal{T}_{\mathrm{enc}}}
\newcommand{\ta}{\mathcal{T}_{a}}
\newcommand{\rbeta}{\mathcal{R}_{\beta}}
\newcommand{\rquart}{\mathcal{R}_{\beta/4}}
\newcommand{\tp}{\mathit{top}}
\newcommand{\ev}{\mathit{ev}}
\newcommand{\bd}{\mathit{bd}}
\newcommand{\sub}{\mathit{sub}}
\newcommand{\rz}{\mathit{rz}}
\newcommand{\lt}{\mathit{lt}}
\newcommand{\sg}{\mathit{sg}}
\newcommand{\inv}{\mathit{inv}}

\newcommand{\phead}[1]{\par\smallskip\noindent\emph{#1.}\ }

%% file: theory.tex
\section{Preliminaries}\label{sec:prelim}

$[n]=\{0,\dots,n-1\}$. $\Fp$ is a prime field, $\K\supseteq\Fp$ a finite extension from which all
verifier challenges are drawn. Integers are identified with their residues when the context is
$\Fp$; for $x\in\Fp$ we write $\langle x\rangle$ for its representative in $(-p/2,p/2)$. For a
predicate $\pi$, $[\pi]\in\{0,1\}$ is its indicator. All matrix dimensions are powers of two
(operands are padded with zeros), and $\nu_m=\log_2 m$, etc.

A function $f:\{0,1\}^{\nu}\to\K$ has a unique multilinear extension
$\tilde f(r)=\sum_{x\in\{0,1\}^\nu}\eq(r,x)f(x)$ with
$\eq(r,x)=\prod_{i}(r_ix_i+(1-r_i)(1-x_i))$. A matrix $F\in\K^{m\times n}$ is a function on
$\{0,1\}^{\nu_m+\nu_n}$ and $\tilde F(r_I,r_J)$ denotes its extension. A non-zero multilinear
polynomial in $\nu$ variables vanishes at a uniform point of $\K^{\nu}$ with probability at most
$\nu/|\K|$~\cite{schwartz80}.

\subsection{Floating-Point Formats and Rounding}\label{sec:formats}

We work with binary formats with an 8-bit significand and a configurable exponent range.
\Cref{rem:precision} discusses other precisions.

\begin{definition}[Format]\label{def:format}
For integers $e_0$ and $H\ge1$,
\[
 \Fl(e_0,H)=\bigl\{\sigma M 2^{Q+e_0}:\sigma\in\{\pm1\}, M\in[256], Q\in[H]\bigr\}.
\]
A triple $(\sigma,M,Q)$ is \emph{canonical} if $M\ge128$ or $Q=0$, and $\sigma=+1$ when $M=0$. The
\emph{unit in the last place} of a canonical triple is $u=2^{Q+e_0}$, and
$\Omega=(255+\tfrac12)\,2^{H-1+e_0}$ is the \emph{overflow threshold}.
\end{definition}

Every element of $\Fl(e_0,H)$ has exactly one canonical triple. The finite \bfs{} numbers are
$\Fl(-133,254)$ (normal numbers $(128+f)2^{e-134}$, subnormals $f\cdot2^{-133}$); formats with $H<254$
are \emph{windowed} \bfs{} with gradual underflow at $2^{e_0}$. Adjacent elements have canonical
significands of opposite parity.

\begin{definition}[Rounding]
For real $v$ with $|v|<\Omega$, $\RN_{e_0,H}(v)$ is the element of $\Fl(e_0,H)$ nearest to $v$; if
two elements are equally near, it is the one whose canonical significand is even. For
$|v|\ge\Omega$ the rounding \emph{overflows} and $\RN$ is undefined.
\end{definition}

\begin{lemma}[Characterization of correct rounding]\label{lem:round}
Let $c\in\Fl(e_0,H)$ have canonical triple $(\sigma,M,Q)$ and unit $u=2^{Q+e_0}$, let $v\in\R$, and
put $\rho=\sigma(v-c)$ and
\[
 \lambda=\begin{cases} u/4 & \text{if } M=128 \text{ and } Q\ge1,\\ u/2 & \text{otherwise.}\end{cases}
\]
Then $|v|<\Omega$ and $c=\RN_{e_0,H}(v)$ if and only if
\begin{enumerate}[label=(\alph*),leftmargin=2em,itemsep=0pt,topsep=2pt]
\item $-\lambda\le\rho\le u/2$, and
\item if $M$ is odd, both inequalities in (a) are strict.
\end{enumerate}
\end{lemma}
\begin{proof}[Proof outline]Conditions (a)--(b) state that $v$ lies in the rounding interval of $c$, whose lower half-width is $u/4$ exactly at a binade boundary and whose endpoints belong to $c$ iff ties go to $c$; the proof checks $c=0$, normal outputs, the binade boundary, the largest value and gradual underflow. The full proof is in Section~\ref{sup:prf:lem:round} of the supplementary material.\end{proof}

\subsection{Interactive Oracle Proofs and Standard Subprotocols}\label{sec:tools}

A \emph{polynomial IOP} sends oracles to multilinear polynomials, queried by the verifier at points
of $\K^\nu$, and is compiled into an argument with a polynomial commitment and
Fiat--Shamir~\cite{bcs16}. We use three standard subprotocols.

\begin{lemma}[Sumcheck~\cite{lfkn92}]\label{lem:sumcheck}
Let $g\in\K[x_1,\dots,x_\nu]$ have degree at most $\delta$ in each variable. The sumcheck protocol
for the claim $\sum_{x\in\{0,1\}^\nu}g(x)=s$ has $\nu$ rounds; if the claim is false, the verifier
accepts with probability at most $\nu\delta/|\K|$. At the end the verifier needs the value $g(r)$
at one uniformly random point $r\in\K^\nu$.
\end{lemma}

\begin{lemma}[Zero-check]\label{lem:zerocheck}
Let $g_1,\dots,g_q$ be polynomials of individual degree at most $\delta$ in oracle polynomials on
$\{0,1\}^\nu$. To check that every $g_c$ vanishes on $\{0,1\}^{\nu}$, the verifier samples
$\alpha\in\K$ and $r\in\K^\nu$ and the parties run sumcheck on
$\sum_x \eq(r,x)\sum_c\alpha^{c}g_c(x)=0$. If some $g_c$ does not vanish on the cube, the verifier
accepts with probability at most $\bigl(q-1+\nu(\delta+2)\bigr)/|\K|$.
\end{lemma}
\begin{proof}
Batching by $\alpha$ fails with probability $(q-1)/|\K|$, the random $r$ with probability $\nu/|\K|$, and
then \cref{lem:sumcheck} applies with degree $\delta+1$.
\end{proof}

\begin{lemma}[Lookup via logarithmic derivatives~\cite{haboeck22}]\label{lem:logup}
Let $f_1,\dots,f_N\in\K$ and let $t_1,\dots,t_T\in\K$ be distinct, with the characteristic of $\K$
larger than $N$. Then there are $m_1,\dots,m_T\in\Fp$ with
$\sum_{i}\frac{1}{x-f_i}=\sum_{j}\frac{m_j}{x-t_j}$ in $\K(x)$ if and only if
$\{f_i\}\subseteq\{t_j\}$. Consequently, if some $f_i$ is not in the table, then for every choice of
$m$ the identity holds at a uniform $\xi\in\K$ with probability at most $(N+T)/|\K|$.
\end{lemma}

The probability bound of \cref{lem:logup} holds for multiplicities that are fixed before $\xi$ is
sampled. This order is essential. If the multiplicities may depend on $\xi$, the identity can be
forged: for $\xi\notin\Fp$ and distinct $t_0,t_1\in\Fp$, the elements $(\xi-t_0)^{-1}$ and
$(\xi-t_1)^{-1}$ are linearly independent over $\Fp$ (otherwise their ratio, and with it $\xi$, would
lie in $\Fp$); since $[\K:\Fp]=2$ for the quadratic extension used in this paper, they span $\K$ (in
an extension of higher degree their span is a proper subspace, and the forgery needs more table
rows), and for every $f\in\Fp\setminus\{t_0,t_1\}$ there are
$m_0,m_1\in\Fp$ with $(\xi-f)^{-1}=m_0(\xi-t_0)^{-1}+m_1(\xi-t_1)^{-1}$. In every protocol of this
paper the multiplicities are therefore committed, or bound into the Fiat--Shamir transcript, before
the lookup challenges $\gamma$ and $\xi$ are derived. We also require that the number of lookups of
every lookup instance is less than $p$, as the lemma assumes.

Tuples are compressed by powers of a challenge $\gamma$; a tuple of width $c$ outside a table of $T$ rows
collides with a compressed row with probability at most $(c-1)T/|\K|$, and table rows collide with
probability at most $(c-1)T^2/|\K|$. The identity is checked with committed helpers
$h_i=(\xi-f_i)^{-1}$~\cite{haboeck22}, committed as two base-field tables, or with a layered
sumcheck~\cite{papini23}. For $N<p$ lookups the total lookup error $\varepsilon_{\mathrm{lk}}$ is at most
$\bigl((c-1)(T+T^2)+N+T\bigr)/|\K|$ plus the error of the sumcheck for the fractional sums.

\section{Canonical Semantics and Native Kernels}\label{sec:kernels}

Fix formats $\Fl_A=\Fl(e_A,H_A)$, $\Fl_B=\Fl(e_B,H_B)$, $\Fl_C=\Fl(e_C,H_C)$ and, when a bias is
present, $\Fl_b=\Fl(e_b,H_b)$.

\begin{definition}[Correctly rounded GEMM]\label{def:cr}
For $A\in\Fl_A^{m\times k}$, $B\in\Fl_B^{k\times n}$ and $b\in\Fl_b^{n}$ (or $b=0$), let
$v_{ij}=\sum_{l\in[k]}A_{il}B_{lj}+b_j\in\R$. The relation $\mathcal{R}_{\CR}$ consists of all
$(A,B,b,C)$ with $C\in\Fl_C^{m\times n}$ such that $|v_{ij}|<\Omega_C$ and
$C_{ij}=\RN_{e_C,H_C}(v_{ij})$ for all $i,j$.
\end{definition}

The bias is \emph{fused}: it is added to the exact sum before the single rounding. Each $C_{ij}$ is
a function of the multiset of products, thus $\mathcal{R}_{\CR}$ is independent of any order of
evaluation. Every product $A_{il}B_{lj}$ has a significand of at most 16 bits; it is exactly representable
in a binary format with at least 16 significand bits whose exponent range contains it. Binary64
contains all products of finite \bfs{} numbers; binary32 does not (for example
$2^{-133}\cdot2^{-133}=2^{-266}$ is below its range, and products of large \bfs{} numbers overflow
it), hence statements about binary32 kernels below assume that all products lie in its range.

\begin{definition}[Accumulating kernel]
An \emph{accumulating kernel} with unit roundoff $\mathsf{u}=2^{-t}$, $t\ge16$, computes for every
$(i,j)$ a value $\hat s_{ij}$ by adding the $k$ products and the bias in some order and with some
bracketing, each addition rounded to nearest in a binary format with $t$ significand bits, and
outputs $\RN_{e_C,H_C}(\hat s_{ij})$. We assume that every product is exact (it lies in the range of
the accumulation format and no fused multiply--add rounds it differently) and that no addition
overflows or underflows.
\end{definition}

Binary64 accumulation ($t=53$) is of this form for all finite \bfs{} operands and every order;
binary32 accumulation ($t=24$) is when all products and partial sums are normal binary32 numbers, as
for the windowed formats of our proofs (product exponents in $[-96,64]$).

\begin{proposition}[Order-agnostic error bound]\label{prop:higham}
For an accumulating kernel with $k\mathsf{u}<1$,
\[
\begin{gathered}
 |\hat s_{ij}-v_{ij}|\le\beta_{ij}:=\gamma_k\Bigl(\textstyle\sum_{l}|A_{il}||B_{lj}|+|b_j|\Bigr),\\
 \gamma_k=\frac{k\mathsf{u}}{1-k\mathsf{u}}.
\end{gathered}
\]
\end{proposition}
\begin{proof}
The products are exact, hence $\hat s_{ij}$ is the result of recursive summation, in some order, of
$k+1$ real numbers with $k$ rounded additions. For any order of summation of $n$ numbers
$x_1,\dots,x_n$ the computed sum $\hat s$ satisfies
$|\hat s-\sum x_i|\le\gamma_{n-1}\sum|x_i|$~\cite[\S4.2]{higham02}.
\end{proof}

\begin{proposition}[Determinacy]\label{prop:determined}
Let $\hat s$ and $v$ be reals and $\beta'\ge|\hat s-v|$, with $|\hat s|+\beta'<\Omega_C$. If
$\RN(\hat s-\beta')=\RN(\hat s+\beta')$, then $\RN(\hat s)=\RN(v)$. In particular the output of an
accumulating kernel at $(i,j)$ equals $\CR(A,B,b)_{ij}$ unless $v_{ij}$ is within $2\beta_{ij}$ of
a rounding breakpoint of $\Fl_C$.
\end{proposition}
\begin{proof}
$\RN$ is non-decreasing, and both $\hat s$ and $v$ lie in $[\hat s-\beta',\hat s+\beta']$, on which
$\RN$ is constant by assumption. For the second claim, take $\beta'=\beta_{ij}$: if no breakpoint
lies in $[\hat s-\beta_{ij},\hat s+\beta_{ij}]\subseteq[v-2\beta_{ij},v+2\beta_{ij}]$, the
hypothesis holds.
\end{proof}

\Cref{prop:determined} is a sufficient test only. When it fails, the native and the canonical
outputs need not be adjacent, because under cancellation $\beta_{ij}$ can span many intervals of the
target format. For $A=(2^{30},1,-2^{30},0)$ and $B=(1,1,1,1)^{\mathsf T}$, left-to-right binary32
accumulation rounds $2^{30}+1$ to $2^{30}$ and returns $0$, whereas $\CR(A,B)=1$. In particular,
native accumulation is not faithful in general.

\begin{proposition}[No kernel-agnostic single-valued relation]\label{prop:orders}
There are $A\in\Fl(-133,254)^{1\times4}$ and $B\in\Fl(-133,254)^{4\times1}$ for which two
accumulating kernels with binary32 accumulation return different outputs. Consequently any
relation that is satisfied by the output of every accumulating kernel admits more than one output
for some input, and the canonical relation $\mathcal{R}_{\CR}$ is not satisfied by every kernel.
\end{proposition}

\noindent (Proof in Section~\ref{sup:prf:prop:orders} of the supplementary material.) A relation is
therefore tied to one order, multi-valued, or canonical in the sense of \cref{def:cr}.

\Cref{prop:determined} gives a Ziv-style algorithm~\cite{ziv91}: compute $\hat s=AB+b$ with any
accumulating kernel and a bound $\hat\beta\ge\beta$ from $|A||B|+|b|$, output $\RN(\hat s)$ where
$\RN(\hat s-\hat\beta)=\RN(\hat s+\hat\beta)$, and recompute the remaining \emph{flagged} entries exactly.
With binary64 accumulation fewer than $2\cdot10^{-4}$ of the entries of every GPT-2 product are flagged.

Trace arithmetizations of floating-point addition~\cite{garg22,ernstberger24,peigne26} commit to all
$k-1$ intermediate roundings per output, $\Theta(mnk)$ data for a matrix product; we claim no lower
bound for proving a sequentially rounding kernel.

\section{Exact Products by Exponent Bucketing}\label{sec:buckets}

Fix a bucket width $w\ge2$ and $\beta=2^{w}$. We assume that $w$ divides
$H_A,H_B,H_C,H_b$ and write $G_\bullet=H_\bullet/w$ for the number of buckets of each format. We
also assume
\begin{equation}\label{eq:align}
 e_C-e_A-e_B=wd,\ d\ge1;\qquad e_b-e_A-e_B=wd_b,\ d_b\ge0 .
\end{equation}
For windowed formats the offsets $e_\bullet$ are free parameters and are chosen as multiples of $w$,
thus \eqref{eq:align} is a mild condition; \cref{rem:general} treats the general case.

Let $x\in\Fl(e_0,H)$ have canonical triple $(\sigma,M,Q)$, and write
$Q=wg+r$ with $g\in[G]$ and $r\in[w]$. The \emph{limb} of $x$ is $X=\sigma M2^{r}\in\Z$ and its
\emph{bucket} is $g$. Thus
\[
\begin{gathered}
 x=\iota(x)\,2^{e_0}\quad\text{with}\quad \iota(x)=X\beta^{g}\in\Z,\\
 |X|\le\Xmax:=255\cdot2^{w-1}.
\end{gathered}
\]
A tensor is committed as the limb matrix $X$ and the one-hot bucket indicators
$E_g=[\,\text{bucket}=g\,]$, $g\in[G]$. Conversely, any pair $(X,g)$ with
$X\in\mathcal{L}:=\{\sigma M2^r\}$ and $g\in[G]$ is the encoding of the element $X\beta^g2^{e_0}$ of
$\Fl(e_0,H)$, whether or not the triple is canonical.

For $(A,B,b)$ as in \cref{def:cr}, with limbs $X^A,X^B,X^b$ and
buckets $g^A,g^B,g^b$, define for $t\in[T_D]$, $T_D:=G_A+G_B-1$,
\begin{equation}\label{eq:D}
 D^{(t)}_{ij}=\sum_{l\in[k]}\ \sum_{g+h=t} X^A_{il}\,[g^A_{il}=g]\;X^B_{lj}\,[g^B_{lj}=h]\ \in\Z ,
\end{equation}
and
\begin{equation}\label{eq:N}
 N_{ij}=\sum_{t\in[T_D]}D^{(t)}_{ij}\beta^{t}+X^b_j\beta^{g^b_j+d_b} .
\end{equation}

\begin{lemma}[Bucket identity]\label{lem:bucket}
$v_{ij}=N_{ij}\,2^{e_A+e_B}$ for all $i,j$. Moreover
$|D^{(t)}_{ij}|\le k\Xmax^{2}$, and for the matrix polynomials
$A(z)=\bigl(X^A_{il}z^{g^A_{il}}\bigr)$ and $B(z)=\bigl(X^B_{lj}z^{g^B_{lj}}\bigr)$ over $\Z[z]$,
\[
 A(z)\,B(z)=\sum_{t\in[T_D]}D^{(t)}z^{t}.
\]
\end{lemma}
\begin{proof}
$A_{il}B_{lj}=X^A_{il}X^B_{lj}\beta^{g^A_{il}+g^B_{lj}}2^{e_A+e_B}$, and grouping the terms by
$t=g^A_{il}+g^B_{lj}$ gives $\sum_lA_{il}B_{lj}=2^{e_A+e_B}\sum_tD^{(t)}_{ij}\beta^t$. By
\eqref{eq:align}, $b_j=X^b_j\beta^{g^b_j}2^{e_b}=X^b_j\beta^{g^b_j+d_b}2^{e_A+e_B}$. Each index $l$
contributes to exactly one $t$, which gives the bound. The polynomial identity is the same grouping
with $z$ in place of $\beta$.
\end{proof}

The verifier cannot check the identity at $z=\beta$ in a small field, because $N_{ij}$ is far larger
than $p$. It checks it at a random $z$ instead, which determines every coefficient separately.

\begin{lemma}[Product check]\label{lem:product}
Let $p>2k\Xmax^2$. Let $X^A,E^A_g$ and $X^B,E^B_g$ be oracles that encode matrices $A,B$ as above,
and let $\hat D^{(t)}$, $t\in[T_D]$, be arbitrary oracles $\{0,1\}^{\nu_m+\nu_n}\to\Fp$. Consider
the protocol: the verifier samples $z\in\K$, $\rho_I\in\K^{\nu_m}$, $\rho_J\in\K^{\nu_n}$; the
parties run the sumcheck of \cite{thaler13} for
\begin{equation}\label{eq:thaler}
\begin{aligned}
 \sum_{t\in[T_D]}z^t\,\tilde{\hat D}^{(t)}(\rho_I,\rho_J)&=\sum_{l\in\{0,1\}^{\nu_k}}\tilde A_z(\rho_I,l)\,\tilde B_z(l,\rho_J),\\
 A_z&=X^A\circ\textstyle\sum_gz^gE^A_g,
\end{aligned}
\end{equation}
($B_z$ analogously, $\circ$ entrywise), which ends with a point $\rho_L$ and prover-supplied values
$a,b'$ whose product must equal the final sumcheck value; the claims
$a=\tilde A_z(\rho_I,\rho_L)$ and $b'=\tilde B_z(\rho_L,\rho_J)$ are each reduced by a sumcheck of
the cubic polynomial $\eq(\cdot,x)\,X(x)\sum_gz^gE_g(x)$ to queries of the oracles $X,E_g$ at one
point. If $\hat D^{(t)}_{ij}\ne D^{(t)}_{ij}\bmod p$ for some $(t,i,j)$, the verifier accepts with
probability at most
\[
 \varepsilon_{\mathrm{prod}}=\frac{T_D-1+4\nu_m+4\nu_n+8\nu_k}{|\K|}.
\]
If the verifier accepts otherwise, $\langle\hat D^{(t)}_{ij}\rangle=D^{(t)}_{ij}$ for all $(t,i,j)$.
\end{lemma}
\begin{proof}[Proof outline]On the cube the bucket polynomials multiply to $\sum_tD^{(t)}z^t$; a false claim survives the random $z$ by Schwartz--Zippel and the sumcheck by \cref{lem:sumcheck}. The full proof is in Section~\ref{sup:prf:lem:product} of the supplementary material.\end{proof}

\section{The Rounding Gadget}\label{sec:gadget}

Fix an output entry $(i,j)$ and drop the indices. We are given the exact integer $N$ of
\eqref{eq:N}, and a claimed output $c\in\Fl_C$, and must enforce $c=\RN(N2^{e_A+e_B})$.

Let $c$ have canonical triple $(\sigma,M,Q)$ with $Q=wg+r$, limb $X=\sigma M2^r$,
and put $P=2^r$ and $\tau=g+d$. By \eqref{eq:align},
\[
\begin{gathered}
 c=X\beta^{\tau}\,2^{e_A+e_B},\qquad u=2^{Q+e_C}=U\,2^{e_A+e_B},\\ U=P\beta^{\tau}.
\end{gathered}
\]
Hence in units of $2^{e_A+e_B}$ the output is $X\beta^\tau$ and its unit in the last place is $U$. Since
$\tau\ge1$ and $w\ge2$, both $U/2$ and $U/4$ are integers.

Put $V=\sigma N-(M-\tfrac12)U\in\Z$, such that $\rho=\sigma(v-c)$ of
\cref{lem:round} is $(V-U/2)\,2^{e_A+e_B}$. The conditions of \cref{lem:round} become
\begin{equation}\label{eq:Vcond}
\begin{gathered}
 0\le V\le U;\qquad V\ge U/4 \ \text{ if } M=128,\ Q\ge1;\\ V\notin\{0,U\}\ \text{ if } M \text{ is odd}.
\end{gathered}
\end{equation}

For each output entry the prover commits to three groups of values. The first is the encoding of
$c$: the limb $X$, the one-hot bucket indicators $E_g$, $g\in[G_C]$, and the attributes $P=2^r$,
$\sg=\sigma$, $\ev=[M\text{ even}]$, $\bd=[M=128]$, $\sub=[M<128]$, $\rz=[r=0]$. The second is a
decomposition of $V$: a bit $\tp$, an integer $a$, a bit $\lt$, an integer $b'$, digits $b_t$ for
$t\in[T_b]$, $T_b:=d+G_C-2$, and a field element $\inv$, with the intended meaning
\begin{equation}\label{eq:V}
 V=\tp\cdot U+a\,\tfrac{\beta^{\tau}}{4}+b'\beta^{\tau-1}+\sum_{t<\tau-1}b_t\beta^{t}.
\end{equation}
The third are the carries: for $t\in[T_c-1]$, where $T_c:=\max(T_D,\,d+G_C,\,d_b+G_b)$, $n_c$ digits
$c_{t,u}\in[\beta]$ that represent the signed carry $c_t=\sum_{u\in[n_c]}c_{t,u}\beta^u-\beta^{n_c}/2$.
The entry also sees the limbs $D^{(t)}$ of \eqref{eq:D} and, if a bias is present, the bias columns
$X^b$ and $E^b_g$, which depend on $j$ only.

The values of each entry are subject to the following lookups.
\begin{enumerate}[label=(L\arabic*),leftmargin=2.6em,itemsep=0pt,topsep=2pt]
\item $(X,P,\sg,\ev,\bd,\sub,\rz)\in\tenc$, the table of all such tuples for $\sigma\in\{\pm1\}$ and
either $M\in[128,256)$, $r\in[w]$, or $M\in[128]$, $r=0$ (with $\sigma=+1$ for $M=0$). It has
$256w+255$ rows, and its rows are in bijection with their first components.
\item $(a,P,\lt)\in\ta=\{(a,2^r,[a<2^r]) : r\in[w],\ a\in[4\cdot2^r]\}$.
\item $b'\in\rquart=[\beta/4]$.
\item $b_t\in\rbeta=[\beta]$ for all $t\in[T_b]$, and $c_{t,u}\in\rbeta$ for all $t\in[T_c-1]$,
$u\in[n_c]$.
\end{enumerate}

The constraints are as follows, with $E_g:=0$ for $g\notin[G_C]$, $b_t:=0$ for $t\ge T_b$, $D^{(t)}:=0$ for
$t\ge T_D$, $c_{-1}:=c_{T_c-1}:=0$ and $S:=a+b'+\sum_{t\in[T_b]}b_t$:
\begin{enumerate}[label=(C\arabic*),leftmargin=2.6em,itemsep=0pt,topsep=2pt]
\item $E_g(1-E_g)=0$ for $g\in[G_C]$, and $\sum_gE_g=1$.
\item $\sub\cdot(1-E_0)=0$.
\item $\tp(1-\tp)=0$,\quad $\tp\cdot a=0$,\quad $\tp\cdot b'=0$,\quad $\tp\cdot b_t=0$ for $t\in[T_b]$.
\item $b_t\cdot\sum_{g\le t-d+1}E_g=0$ for $t\in[T_b]$.
\item $\bd\cdot(1-E_0\,\rz)\cdot(1-\tp)\cdot\lt=0$.
\item $(1-\ev)\cdot\tp=0$ and $(1-\ev)\cdot(S\cdot\inv-1)=0$.
\item For every $t\in[T_c]$: \ $e_t+c_{t-1}-\beta c_t=0$, where
\begin{align*}
 e_t=\ & 2D^{(t)}+2X^b E^b_{t-d_b}-(2X-\sg\,P)\,E_{t-d}\\
       & -2\,\sg\Bigl(\tp\,P\,E_{t-d}+\bigl(a\tfrac{\beta}{4}+b'\bigr)E_{t-d+1}+b_t\Bigr).
\end{align*}
\end{enumerate}
All constraints have total degree at most $5$ in the columns. Let
\[
 \Emax:=2k\Xmax^2+4\Xmax+2\beta^2+\beta ,
\]
which bounds $|e_t|$ for all column values allowed by (L1)--(L4), (C1), (C3) and true limbs
$D^{(t)}$.

\begin{lemma}[Rounding gadget]\label{lem:gadget}
Assume
\begin{equation}\label{eq:pc}
 \Emax+(\beta+1)\frac{\beta^{n_c}}{2}<p \qquad\text{and}\qquad \frac{\Emax}{\beta-1}<\frac{\beta^{n_c}}{2},
\end{equation}
and, for completeness,
\begin{equation}\label{eq:inv}
 (T_b+3)\,\beta<p\qquad\text{and}\qquad G_C<p .
\end{equation}
Let $N$ be the integer of \eqref{eq:N} for well-formed operand encodings, and suppose the columns
$D^{(t)}$ of the entry hold the residues of the true limbs.
\begin{enumerate}
\item \emph{(Soundness.)} If the columns of the entry satisfy \textup{(L1)--(L4)} and
\textup{(C1)--(C7)} over $\Fp$, then $(X,E)$ is the canonical encoding of an element
$c\in\Fl_C$, $|N2^{e_A+e_B}|<\Omega_C$, and $c=\RN_{e_C,H_C}(N2^{e_A+e_B})$.
\item \emph{(Completeness.)} If $|N2^{e_A+e_B}|<\Omega_C$ and $c=\RN_{e_C,H_C}(N2^{e_A+e_B})$, there
is an assignment of the columns, computable in $O(T_c)$ integer operations, that satisfies
\textup{(L1)--(L4)} and \textup{(C1)--(C7)}.
\end{enumerate}
\end{lemma}
\begin{proof}[Proof outline]Soundness decodes the limbs from the lookups, lifts every field equation to the integers by \eqref{eq:inv} and the limb bounds, telescopes the carry chain to the residual of \cref{lem:round}, and applies that lemma; completeness exhibits the honest witness and uses \eqref{eq:inv} for the inverse. The full proof is in Section~\ref{sup:prf:lem:gadget} of the supplementary material.\end{proof}

\section{The Protocol}\label{sec:protocol}

The statement consists of commitments to the encodings of $A$, $B$, $b$ and
$C$: the limb matrix and the bucket indicators of each. The first prover message contains, as
oracles on $\{0,1\}^{\nu_m+\nu_n}$, the attribute, decomposition and carry columns of
\cref{sec:gadget} and the limbs $\hat D^{(t)}$. The bias columns of an output row are the bias
oracles evaluated on the $j$-coordinates.

The protocol $\Pi_{\CR}$: (1) the prover sends these oracles; (2) it sends or binds the lookup
multiplicities, and only then does the verifier send $\gamma,\xi$, after which the prover sends the
helper oracles; (3) the product check of \cref{lem:product}; (4) the zero-check of (C1)--(C7) and the
fractional-sum checks for (L1)--(L4); (5) unless the operands are outputs of a verified instance or
certified commitments, a zero-check that their indicators are one-hot and a lookup of their limbs in
$\mathcal L$. All oracles are queried at $O(1)$ points.

\begin{theorem}\label{thm:main}
Assume \eqref{eq:align}, \eqref{eq:pc}, \eqref{eq:inv}, $p>2k\Xmax^2$, and that the number of lookups
and of table rows is less than $p$. Then $\Pi_{\CR}$ is a public-coin
polynomial IOP with the following properties.
\begin{enumerate}
\item \emph{Completeness.} If $(A,B,b,C)\in\mathcal{R}_{\CR}$ and the encodings are canonical, the
honest prover makes the verifier accept with probability $1$ (up to the negligible probability
that a lookup challenge hits a pole).
\item \emph{Soundness.} If the oracles of the statement do not encode a tuple
$(A,B,b,C)\in\mathcal{R}_{\CR}$, with $C$ canonically encoded, then for every prover the verifier
accepts with probability at most
\[
 \varepsilon=\varepsilon_{\mathrm{prod}}+\frac{q-1+7(\nu_m+\nu_n)}{|\K|}+\varepsilon_{\mathrm{lk}}+\varepsilon_{\mathrm{op}},
\]
where $q$ is the number of constraints of the output table, $\varepsilon_{\mathrm{prod}}$ is as in
\cref{lem:product}, and $\varepsilon_{\mathrm{op}}$ is the error of the operand check, at most
$\sum_{\Theta\in\{A,B,b\}}\bigl(G_\Theta+1+4\nu_\Theta\bigr)/|\K|$ plus its lookup error, with
$\nu_\Theta$ the number of variables of the operand.
\item \emph{Efficiency.} Beyond the operand encodings the prover commits to
\begin{equation}\label{eq:cells}
 n_{\mathrm{cell}}=(7+G_C)+5+T_b+T_D+n_c(T_c-1)
\end{equation}
field elements and performs $3+T_b+n_c(T_c-1)$ lookups per output entry, in time one exact integer
matrix product plus $O\bigl(G_Amk+G_Bkn+mn\,n_{\mathrm{cell}}\bigr)$ field operations; the verifier performs
$O(\nu_m+\nu_k+\nu_n+n_{\mathrm{cell}}+G_A+G_B+G_b)$ operations plus work linear in the tables.
\end{enumerate}
\end{theorem}
\begin{proof}[Proof outline]Composition of \cref{lem:product,lem:gadget,lem:logup} with a union bound over the verifier's challenges. The full proof is in Section~\ref{sup:prf:thm:main} of the supplementary material.\end{proof}

\begin{corollary}[Succinct argument]\label{cor:arg}
If $\Pi_{\CR}$ is compiled with an extractable multilinear commitment and the Fiat--Shamir
transformation, and the hypotheses of the compilation theorem of~\cite{bcs16} hold, the result is a
non-interactive argument of knowledge for $\mathcal{R}_{\CR}$ in the random-oracle model. With a
commitment whose openings are polylogarithmic~\cite{fri18,basefold24,whir25,binius23}, proof size and
verification time are polylogarithmic in $mnk$ for fixed $w$, formats and tables; our
instantiation has $O(\sqrt N)$ openings and does not achieve this.
\end{corollary}

\label{par:inst}
\emph{Instantiation.} Our native prover (\cref{sec:native}) uses $p=2^{64}-2^{32}+1$,
$\K=\Fp[x]/(x^2-7)$ ($|\K|\approx2^{128}$), and a Ligero--Brakedown commitment~\cite{ligero17,brakedown23}:
evaluation tables are rows of a matrix, Reed--Solomon encoded at rate $1/4$, with BLAKE3 Merkle trees
over columns; an opening sends the row combinations for the point and for a random combination and
checks them against $250$ columns, with error about $(1-\delta/3)^{250}<2^{-103}$ for $\delta=3/4$ plus
terms of order (rows)$/|\K|$~\cite{brakedown23}. This commitment is not hiding. Lookup multiplicities are
absorbed into the transcript before $\gamma,\xi$; parameters are committed and certified once in a
setup phase. For $N\approx2^{28}$ lookups, $T<2^{15}$ and $c=7$ the lookup error is below $2^{-90}$.

\begin{remark}[Security of the compiled argument]\label{rem:zk}
All bounds of this paper are interactive soundness errors. A bound for the Fiat--Shamir argument must
be derived for the complete message schedule (helper commitments, delayed openings), which we do
not do; the argument is transparent and plausibly post-quantum, and we claim no quantum security
level~\cite{cms19}. Remarks on general exponent alignment (full \bfs{} with $H=254$ is not an instance of
\cref{thm:main} as stated) and on other precisions are in Section~\ref{sup:ext} of the supplementary
material.
\end{remark}

\emph{Extensions.} A
faithful-rounding variant, bit-exact proofs of native kernels, lattice and binary-field commitments,
and the leakage of windowed formats are discussed in Section~\ref{sup:ext} of the supplementary
material. RMSNorm, rotary embeddings and gated activations need only the ingredients of \cref{sec:elem}.

\section{Element-Wise Layers and an End-to-End Block}\label{sec:elem}

Every operation of a transformer block is an exact computation followed by one correct rounding
into the format of its output, and each tensor is committed once.

\phead{Sums, Differences and Products}\label{sec:elem-arith}

Let $a\in\Fl(e_a,H_a)$ and $b\in\Fl(e_b,H_b)$ have \emph{valid} encodings $(X_a,E^a)$ and $(X_b,E^b)$
in the sense of \cref{sec:buckets} (a limb in $\mathcal L$ and one-hot indicators, not necessarily
canonical), with $G_a=H_a/w$ and $G_b=H_b/w$ buckets, and let the output format be $\Fl(e_c,H_c)$.
The lemmas below use the operand columns only through the products $X\,E_q$, $q\in[G]$; they
therefore also apply to an \emph{auxiliary factorization} $(X,E_0,\dots,E_{G-1})$ of field elements
whose products satisfy $X E_q=X'[q=g']$ for a valid encoding $(X',g')$, as produced by
\cref{lem:max} below.

\begin{lemma}[Limbs of element-wise operations]\label{lem:elem-int}
\begin{enumerate}
\item \emph{Sum and difference.} Let $u=\min(e_a,e_b,e_c-w)$ and assume that $w$ divides $e_a-u$,
$e_b-u$ and $e_c-u$; put $\delta_a=(e_a-u)/w$, $\delta_b=(e_b-u)/w$ and $d=(e_c-u)/w\ge1$. For
$s\in\{+1,-1\}$ set $D^{(t)}=X_aE^a_{t-\delta_a}+s\,X_bE^b_{t-\delta_b}$ for
$t\in[T_D]$, $T_D=\max(G_a+\delta_a,G_b+\delta_b)$, with $E_g:=0$ outside $[G]$. Then
$N=\sum_tD^{(t)}\beta^t$ satisfies $N2^{u}=a+s\,b$, and $|D^{(t)}|\le2\Xmax$.
\item \emph{Product with fused bias.} Let $u=e_a+e_b$, assume $w\mid(e_c-u)$ with $d=(e_c-u)/w\ge1$,
and, for a bias $\theta\in\Fl(e_\theta,H_\theta)$, $w\mid(e_\theta-u)$ with $d_\theta=(e_\theta-u)/w\ge0$. Set
$D^{(t)}=X_aX_b\sum_{i+j=t}E^a_iE^b_j$ for $t\in[G_a+G_b-1]$. Then
$N=\sum_tD^{(t)}\beta^t+X_\theta\beta^{g_\theta+d_\theta}$ satisfies $N2^u=ab+\theta$, and
$|D^{(t)}|\le\Xmax^2$.
\end{enumerate}
In both cases $c=\RN(N2^u)$ is the quantity enforced by the rounding gadget of \cref{sec:gadget}
with unit $2^{e_A+e_B}:=2^u$.
\end{lemma}
\begin{proof}
Exactly one indicator of a valid encoding is $1$ (for a factorization, the products $XE_q$ vanish
except at $q=g'$), hence
$\sum_tX_aE^a_{t-\delta_a}\beta^t=X_a\beta^{g_a+\delta_a}=\iota(a)\,2^{e_a-u}$, and likewise for $b$.
Hence $N=(a+s\,b)2^{-u}$. At most one term of each operand is non-zero in each position, which
gives the bound $2\Xmax$. In the product case the one-hot property leaves the single term
$X_aX_b\beta^{g_a+g_b}=\iota(a)\iota(b)$, and $ab=\iota(a)\iota(b)2^{e_a+e_b}$; the bias term is that of
\eqref{eq:N}.
\end{proof}

These limbs are polynomials of degree at most $2$ (sums) and $4$ (products) in the operand columns and
are substituted into (C7) instead of being committed; the gadget keeps degree $5$ (6 with the
zero-check weight), and $\Emax$ of \cref{lem:gadget} becomes
\begin{equation}\label{eq:emax-elem}
\begin{gathered}
 \Emax:=2L+4\Xmax+2\beta^2+\beta,\\ L=\begin{cases}2\Xmax&\text{sum, difference},\\ \Xmax^2+\Xmax&\text{product with bias}.\end{cases}
\end{gathered}
\end{equation}
For $w=12$ and $n_c=4$, $\Emax<5.46\cdot10^{11}$ and \eqref{eq:pc} holds for $p=2^{64}-2^{32}+1$.

\begin{lemma}[Element-wise rounding]\label{lem:elem-round}
Assume \eqref{eq:pc} and \eqref{eq:inv} for the $\Emax$ of \eqref{eq:emax-elem}. If the operands
are valid encodings or auxiliary factorizations and the output columns satisfy \textup{(L1)--(L4)} and \textup{(C1)--(C7)} with the limbs
of \cref{lem:elem-int}, then the output is the canonical encoding of $\RN(a\pm b)$, respectively of
$\RN(ab+\theta)$, and that value lies in the output format. Conversely, for valid operands whose
correctly rounded result lies in the output format there is a satisfying assignment.
\end{lemma}
\begin{proof}
The products $X E_q$ of the operand columns are bounded by $\Xmax$ and non-zero for at most one $q$, hence each limb expression
is the residue of an integer of absolute value at most $L<p/2$, and \cref{lem:elem-int} identifies
$N$ and the unit. \Cref{lem:gadget} then applies verbatim with this $\Emax$. The converse is the
completeness half of \cref{lem:gadget}.
\end{proof}

\phead{Unary Maps by Lookup}\label{sec:elem-unary}

A function $\varphi:\R\to\R$ of one variable, an input format $\Fl_{\mathrm{in}}$ and an output
format $\Fl_{\mathrm{out}}$ determine the public table
\[
\begin{aligned}
 \mathcal{T}_\varphi=\;&\bigl\{(1,\mathrm{enc}(x),\mathrm{enc}(\RN_{\mathrm{out}}(\varphi(x)))) : x\in\mathrm{dom}\bigr\}\\
 &\cup\bigl\{(0,\mathrm{enc}(x),\mathrm{enc}(0)) : x\in\Fl_{\mathrm{in}}\bigr\},
\end{aligned}
\]
where $\mathrm{enc}(x)=(X,g)$ is the canonical limb and bucket, $\mathrm{dom}$ is the set of
inputs for which the function is defined and its rounded value lies in $\Fl_{\mathrm{out}}$, and
the second part is present only for masked maps. A row of a unary layer is the tuple
$(\mu_x,X_{\mathrm{in}},g_{\mathrm{in}},X_{\mathrm{out}},g_{\mathrm{out}})$, where $\mu_x\in\{0,1\}$ is a
public mask (the causal mask for the exponential of the softmax, the constant $1$ otherwise) and
$g=\sum_q qE_q$ is read from the one-hot indicators.

\begin{lemma}[Unary lookup]\label{lem:gelu}
Let $N$ rows be compressed by powers of $\gamma$ into $f_1,\dots,f_N$, and let the output indicators
satisfy (C1). If $\sum_i(\xi-f_i)^{-1}=\sum_jm_j/(\xi-t_j)$, where the $t_j$ are the compressed rows of
$\mathcal{T}_\varphi$, then, except with probability $5N|\mathcal{T}_\varphi|/|\K|$ plus the error
of \cref{lem:logup}, every row lies in $\mathcal{T}_\varphi$. A row with $\mu_x=1$ is then a canonical
input in $\mathrm{dom}$ and the canonical encoding of $\RN_{\mathrm{out}}(\varphi(x))$; a row with
$\mu_x=0$ has output $0$.
\end{lemma}
\begin{proof}
A tuple of width $5$ outside the table is compressed by a polynomial of degree $5$ in $\gamma$, thus it
collides with some compressed table row with probability at most $5|\mathcal{T}_\varphi|/|\K|$; the union over the $N$ rows gives the first term, and
\cref{lem:logup} the second. Since the indicators are one-hot, $g=\sum_qqE_q$ is the index of the
active bucket, hence a table row determines the encodings.
\end{proof}

The block uses four maps: GELU in GPT-2's tanh form, $x\mapsto e^{-x/8}$ on $x\ge0$ (masked), $1/x$
and $\sqrt{768/x}$. Each table is published in the plan; it is computed in binary64 and rounded, which
is not a certified correct rounding of the real function, so all semantic statements are relative to
the published tables. A lookup keyed on $(X,g)$ needs a canonical input; every unary map in the block
reads the output of a rounding gadget, which is canonical by \cref{lem:gadget}.

\phead{Row Reductions}\label{sec:elem-rows}

LayerNorm and softmax need row-wise sums $c_i=\RN(\sum_lA_{il}B_{il}+\theta)$ and
$c_i=\RN(\sum_lA_{il})$. These are the gadget of \cref{sec:gadget} with committed limbs
$D^{(t)}_i=\sum_l\sum_{g+h=t}X^A_{il}[g^A_{il}=g]X^B_{il}[g^B_{il}=h]$ (with $X^B\equiv1$, $g^B\equiv0$ for a
plain sum).

\begin{lemma}[Row product check]\label{lem:rowprod}
Let $\hat D^{(t)}$ be arbitrary oracles on $\{0,1\}^{\nu_m}$. The verifier samples $z\in\K$ and
$\rho\in\K^{\nu_m}$, and the parties run sumcheck for
$\sum_tz^t\tilde{\hat D}^{(t)}(\rho)=\sum_{x\in\{0,1\}^{\nu_m+\nu_k}}\eq(\rho,x_I)A_z(x)B_z(x)$, followed
by the reductions of \cref{lem:product} for the final values of $A_z$ and $B_z$. If some
$\hat D^{(t)}_i\ne D^{(t)}_i$, the verifier accepts with probability at most
$\bigl(T_D-1+\nu_m+9(\nu_m+\nu_k)\bigr)/|\K|$.
\end{lemma}
\begin{proof}
As for \cref{lem:product}: a false limb makes $\sum_tz^t(\hat D^{(t)}-D^{(t)})$ a non-zero vector except
with probability $(T_D-1)/|\K|$, its multilinear extension is then non-zero at $\rho$ except with
probability $\nu_m/|\K|$, the summand has individual degree $3$, and each reduction contributes
$3(\nu_m+\nu_k)/|\K|$.
\end{proof}

\phead{LayerNorm and Softmax in Canonical Form}\label{sec:elem-ln}

\begin{definition}[Canonical LayerNorm]\label{def:ln}
For a row $x\in\R^{1024}$ whose first $768$ entries are the activations, gain $\gamma$, offset
$\theta$ and the public dyadic constant $\hat\epsilon=\RN_{\epsilon}(9\cdot768\cdot10^{-5})$ in its
public format, let
\begin{align*}
 z_j&=\RN\bigl(3x_j-2^{-8}\textstyle\sum_lx_l\bigr), & r&=\RN\bigl(\sqrt{768/q}\bigr),\\
 q&=\RN\bigl(\textstyle\sum_jz_j^2+\hat\epsilon\bigr), & t_j&=\RN(z_j\,r),\\
 h_j&=\RN(t_j\gamma_j+\theta_j). &&
\end{align*}
\end{definition}
Here $z_j=0$ for $j\ge768$ and the sum in $q$ runs over all $1024$ positions. The first step is a
matrix product with the public $2048\times1024$ matrix $M$ ($M_{j,j}=3$, $M_{1024+l,j}=-2^{-8}$ for
$j,l<768$, zero otherwise, exact in $\Fl(-48,48)$), whose padded columns are zero; $q$ is a row
product with fused bias, $r$ a lookup, $t,h$ products. Without roundings $t_j=(x_j-\mu)/\sqrt{\sigma^2+\epsilon'}$
with $\epsilon'=\hat\epsilon/(9\cdot768)$; since $9\cdot768\cdot10^{-5}$ is not dyadic, the canonical
semantics uses $\hat\epsilon=71\cdot2^{-10}$, i.e.\ $\epsilon'\approx1.0031\cdot10^{-5}$, by definition.

\begin{definition}[Canonical causal softmax]\label{def:sm}
For a score row $S_{i,\cdot}$ of head $h$ and causal mask $j\le i$, let
$c_i=\max_{j\le i}S_{ij}$, $\Delta_{ij}=\RN(c_i-S_{ij})$,
$E_{ij}=\RN(e^{-\Delta_{ij}/8})$ for $j\le i$ and $E_{ij}=0$ otherwise,
$Z_i=\RN(\sum_jE_{ij})$, $R_i=\RN(1/Z_i)$ and $P_{ij}=\RN(E_{ij}R_i)$.
\end{definition}
The scale $1/\sqrt{64}=1/8$ is absorbed into the exponential table. Without roundings $P$ is the
softmax of $S/8$ over the causal window.

The maximum $c_i$ is not a rounding, and is enforced by a committed selector
$\Sigma_{ij}\in\{0,1\}$:

\begin{lemma}[Maximum by selection]\label{lem:max}
Let $S$ be canonical encodings in $\Fl(e_S,H)$, let $c$ be given by arbitrary oracles $(X^c,E^c_0,\dots,E^c_{G-1})$
whose encoded value is $\bar c_i=\sum_qX^c_iE^c_{i,q}\beta^q2^{e_S}$, and let $\Delta$ be in a format with
$e_\Delta\le e_S$. Write $S^z_{ij}=X^S_{ij}z^{g^S_{ij}}$ and $c^z_i=X^c_i\sum_qE^c_{i,q}z^q$. Suppose (i)
$\Sigma_{ij}(1-\Sigma_{ij})=0$ and $\Sigma_{ij}[j>i]=0$ for all $i,j$; (ii)
$\sum_j\Sigma_{ij}(1+\lambda S^z_{ij})=1+\lambda c^z_i$ for all $i$, for verifier challenges $\lambda,z$; and
(iii) every unmasked row of the exponential lookup is in its table, whose domain is $\Delta\ge0$, with
$\Delta_{ij}=\RN(\bar c_i-S_{ij})$ proved by the difference gadget; and (iv) every row has fewer than $p$
entries. Then, except with probability
$(1+H/w)/|\K|$ per row, $\bar c_i=\max_{j\le i}S_{ij}$, and $|X^c_iE^c_{i,q}|\le\Xmax$ for every $q$.
\end{lemma}
\begin{proof}[Proof outline]The selectors are bits and, by (ii) and (iv), exactly one is set per row; the polynomial identity in $z$ gives the products $X^cE^c_q$; the domain of the exponential lookup gives $\bar c_i\ge S_{ij}$. The full proof is in Section~\ref{sup:prf:lem:max} of the supplementary material.\end{proof}

\phead{Chaining Layers Through Views}\label{sec:chain}

Each tensor is committed once, by the step that produces it; consumers read it in other shapes (a
head, a transpose, concatenated heads, a broadcast scalar) through \emph{views}.

\begin{definition}[View]
A view of $f:\{0,1\}^\nu\to\K$ is $f\circ\sigma$, where $\sigma:\{0,1\}^{\nu'}\to\{0,1\}^\nu$ sets each
output coordinate to a constant bit or to an input coordinate, no input coordinate used twice.
\end{definition}

\begin{lemma}[Views commute with multilinear extension]\label{lem:view}
For the same coordinate map $\hat\sigma$ on field points, $\widetilde{f\circ\sigma}=\tilde f\circ\hat\sigma$.
\end{lemma}
\begin{proof}
$\tilde f\circ\hat\sigma$ is multilinear and agrees with $f\circ\sigma$ on the cube.
\end{proof}

Hence every claim of a consumer is a claim about the producer's polynomial; there are no copies or
copy constraints, all claims on a commitment are discharged by one batched opening, and a consumer
cannot be given a tensor other than the one its producer was proved to output.

\phead{The Block Argument}\label{sec:block}

\begin{table*}[t]
\centering\footnotesize
\caption{The steps of one GPT-2 block. Every tensor has its own public format, is committed once,
and is read by later steps through views. Heads are padded from 12 to 16 and widths from 768 to
1024 and 3072 to 4096; the honest prover sets the padded parameters to zero, but the verifier does
not enforce it (\cref{sec:zk-model}).}\label{tab:steps}
\begin{tabular}{lll}
\toprule
step & relation & argument\\
\midrule
$z_1=\RN(x\,M)$ & matrix product, operand read as $(i,s,l)\mapsto x_{il}$ & \cref{thm:main}\\
$q_1=\RN(\sum_jz_{1,j}^2+\hat\epsilon)$ & row product, fused bias & \cref{lem:rowprod,lem:gadget}\\
$r_1=\RN(\sqrt{d/q_1})$ & unary lookup & \cref{lem:gelu}\\
$t_1=\RN(z_1r_1)$, $h_1=\RN(t_1\gamma_1+\theta_1)$ & products, broadcast views & \cref{lem:elem-round}\\
$\mathit{qkv}=\RN(h_1W_a+b_a)$ & matrix product & \cref{thm:main}\\
$S_h=\RN(Q_hK_h^{\mathsf T})$ & 16 batched products, views of $\mathit{qkv}$ & \cref{thm:main}\\
$c,\Sigma$ & maximum by selection & \cref{lem:max}\\
$\Delta=\RN(c-S)$, $E$, $Z$, $R$, $P$ & difference, masked lookup, row sum, lookup, product & \cref{sec:elem}\\
$O_h=\RN(P_hV_h)$ & 16 batched products & \cref{thm:main}\\
$a=\RN(OW_p+b_p)$, $x_1=\RN(x+a)$ & matrix product (view concatenates heads), sum & \cref{thm:main,lem:elem-round}\\
LayerNorm 2, $f=\RN(h_2W_f+b_f)$ & as above & \\
$g=\RN(\mathrm{GELU}(f))$ & unary lookup & \cref{lem:gelu}\\
$m=\RN(gW_m+b_m)$, $x_2=\RN(x_1+m)$ & matrix product, sum & \cref{thm:main,lem:elem-round}\\
\bottomrule
\end{tabular}
\end{table*}

\begin{definition}[Canonical block]\label{def:block}
Fix public formats for every tensor of \cref{tab:steps}, the public constants $M$ and
$\hat\epsilon$, and the published tables of the unary maps. The canonical block maps an input $x$ and
parameters $(\gamma_1,\theta_1,W_a,b_a,W_p,b_p,\gamma_2,\theta_2,W_f,b_f,W_m,b_m)$ to the output $x_2$
computed by the steps of \cref{tab:steps}, each an exact computation followed by one correct
rounding or by a lookup in the published table (\cref{def:ln,def:sm}). The relation
$\mathcal{R}_{\mathrm{block}}$ contains the commitments to $x$, the parameters and $x_2$ together with
all intermediate tensors when (i) the input and the parameters are valid encodings of elements of
their formats, (ii) every tensor produced by a rounding gadget or a lookup is the canonical encoding
of its defined value, (iii) each maximum $c_i$ is given by an auxiliary factorization whose products
equal those of the selected score, as in \cref{lem:max}, and (iv) no value leaves its format.
\end{definition}

In $\Pi_{\mathrm{block}}$ the prover commits to every tensor and sends the multiplicities of one
global lookup; after $\gamma,\xi$ it commits to the helpers. The parties run, in the order of
\cref{tab:steps}, the product checks and gadget zero-checks of all rounded tensors, the lookup
zero-checks, the certification of input and parameter encodings, and the checks of \cref{lem:max},
reading operands through views; the verifier checks the global lookup identity, and every commitment
is opened once at all its claim points.

\begin{theorem}\label{thm:block}
$\Pi_{\mathrm{block}}$ is complete, and if the committed tensors are not in
$\mathcal{R}_{\mathrm{block}}$, the verifier accepts with probability at most the sum of the soundness
errors of \cref{thm:main} and \cref{lem:rowprod,lem:elem-round,lem:gelu,lem:max} over the steps,
plus the error of the global lookup and of the commitment openings.
\end{theorem}
\begin{proof}[Proof outline]Induction over the steps of \cref{tab:steps}, using the lemmas of this section for every step and a union bound. The full proof is in Section~\ref{sup:prf:thm:block} of the supplementary material.\end{proof}

Blocks compose through their input and output commitments (\cref{sec:zk}).

\input{theory_zk}

%% file: theory_zk.tex
\section{Zero Knowledge and the Complete Model}\label{sec:zk}

The arguments of \cref{sec:protocol,sec:elem} reveal evaluations, partial sums and helper sums. We
now give a privacy construction for the \emph{complete} forward pass, designed so that the verifier
learns only the prompt, the public plan and the last-position logits, and prove its zero knowledge
under explicit hypotheses $\mathsf H$ and $\mathsf E$: a hiding commitment, masked
sumchecks, zero-sum checks, hidden lookup sums, and the complete model, in this order.

\begin{definition}[Model relation]\label{def:rmodel}
A \emph{plan} $\mathcal{P}$ fixes the architecture (GPT-2 small with $L$ blocks, padded as in
\cref{tab:steps}), the views between tensors, the public format of every tensor, the public
constants (among them $\hat\epsilon$ of \cref{def:ln}), and the published tables of every unary map.
For a prompt $\tau\in[V]^T$ and a vector $y$ of $V$ logits, the relation
$\mathcal{R}_{\mathrm{model}}$ contains $((\mathcal{P},\tau,y);\,w)$ when $w$ consists of parameters
$\theta$ (token and position embeddings, the parameters of every block and of the final LayerNorm),
given as valid encodings in their formats, and of all tensors of the canonical forward pass on
$\tau$ with parameters $\theta$ in the sense of \cref{def:block}: every step is an exact
computation followed by one correct rounding, the unary maps are the published tables of
$\mathcal{P}$ (not the real functions), no value leaves its format, and $y$ is the last row of the
logits $h_f\,\mathrm{wte}^{\mathsf T}$, rounded to the logit format.
\end{definition}

The architecture is the padded one of \cref{tab:steps}, and parameters at padded coordinates are
not required to vanish: the relation is that of the widened network, and a prover may use the
padded coordinates as hidden capacity. A public mask $Z$ and a check $Z\circ\theta=0$ inside an
existing zero-check would restrict it to the original network; we have not implemented it. The
parameters are part of the witness, hence a proof states that \emph{some} model of the (widened)
architecture and formats maps the prompt to the logits under the published tables.

\begin{definition}[Honest-verifier zero knowledge]
A public-coin argument is \emph{statistical honest-verifier zero knowledge} with error
$\varepsilon_{\mathrm{zk}}$ if an efficient simulator $\mathcal S(x)$ produces, for every $(x;w)\in\mathcal R$, a
distribution within statistical distance $\varepsilon_{\mathrm{zk}}$ of the honest verifier's view (coins,
prover messages, Merkle roots); the hash function is a random oracle that the simulator answers by
lazy sampling and may program.
\end{definition}

The implementation derives all challenges by Fiat--Shamir from a BLAKE3 transcript; we have not
verified the hypotheses of the compiler theorems~\cite{bcs16} for the complete schedule, and all
security statements concern the interactive protocol.

\phead{A Hiding Polynomial Commitment}\label{sec:zk-pcs}

Let $f_1,\dots,f_{n_p}$ be tables on $\{0,1\}^\nu$ with values in $\Fp$ that will be opened at $K$
points (a number fixed by the plan). A table with values in $\K$, such as a lookup helper, is
committed as two base-field tables, its components in the basis $1,\sqrt7$, and is evaluated by
combining the two evaluations. The prover commits instead to the tables $\hat f_p$ on
$\{0,1\}^{1+\nu}$ with
\[
 \hat f_p(0,x)=f_p(x),\qquad \hat f_p(1,x)\ \text{uniform in }\Fp ,
\]
that is, every table gets a leading \emph{pad variable} whose upper half is a one-time pad. As in
\cref{sec:protocol}, the $n_p2^{1+\nu}$ values are laid out as $n_pR$ rows of length $c$, the row
index being the leading $\log R$ variables. Below them the prover appends $2(K+1)$ uniformly random
\emph{masking rows} $M_0,M_0',\dots,M_K,M_K'\in\Fp^{c}$; we write $M_k^{\K}=M_k+\sqrt7\,M_k'\in\K^c$,
which is uniform in $\K^c$.

\emph{Encoder and queries.} Every row is extended by $e$ random symbols and encoded in coefficient
form, $\bigl(\sum_{s<c+e}m_s\omega^{js}\bigr)_{j\in[n]}$, at rate below $0.266$; the columns are hashed into a
Merkle tree, and an opening queries $Q$ positions uniformly with replacement. At distinct positions
the map from a row's extension symbols to its codeword symbols is a diagonal matrix times a
Vandermonde matrix, of rank $\min(s,e)$; we use $e=256>Q=250$.

\emph{Opening.} To open at points $z_1,\dots,z_K\in\K^{1+\nu}$, write $z_k=(a_k,b_k)$ for the row
and column parts and $\eq(a_k,\cdot)$, $\eq(b_k,\cdot)$ for the corresponding coefficient vectors. The
prover sends the claimed values $v_{k,p}=\hat f_p(z_k)$ (they are sent during the protocol) and the
mask values $\mu_k=\langle\eq(b_k,\cdot),M_k^{\K}\rangle\in\K$. The verifier sends
$\lambda,\vartheta\in\K\setminus\Fp$. The prover sends, over the extended rows,
\begin{align*}
 u_0&=\textstyle\sum_{i}\vartheta^{i}\,\mathrm{row}_i \quad\text{(all rows, masks included)},\\
 u_k&=\textstyle\sum_{p}\lambda^{p}\sum_{r}\eq(a_k,r)\,\mathrm{row}_{p,r}+\lambda^{n_p}M_k^{\K}.
\end{align*}
The verifier checks, for every $k$,
\begin{equation}\label{eq:zkopen}
 \bigl\langle \eq(b_k,\cdot),u_k|_{c}\bigr\rangle=\textstyle\sum_p\lambda^p v_{k,p}+\lambda^{n_p}\mu_k ,
\end{equation}
where $u_k|_c$ are the first $c$ entries, and then opens the $Q$ queried columns and checks their
Merkle paths and that the encoding of every $u$ agrees with the same combination of the opened
column entries.

\begin{lemma}[Binding]\label{lem:zkbind}
Opening a commitment of this construction has the soundness error of the underlying
scheme~\cite{ligero17,brakedown23} for a matrix with $n_pR+2(K+1)$ rows, increased by
$n_p/(|\K|-p)$ per opening.
\end{lemma}
\begin{proof}[Proof outline]The masking rows are rows of the committed matrix, and the mask term has its own power of $\lambda$; a wrong value survives with probability at most $n_p/(|\K|-p)$. The full proof is in Section~\ref{sup:prf:lem:zkbind} of the supplementary material.\end{proof}

We do not re-derive a knowledge extractor; soundness uses only this binding statement.

\begin{lemma}[Hiding openings]\label{lem:zkhide}
Fix the data $f$ and the opening points. Let $z_k^\circ$ be $z_k$ without its pad coordinate $y_k$,
and let $\Phi$ be the $\Fp$-linear map $\pi\mapsto(\tilde\pi(z_k^\circ))_{k\in[K]}$ from $\Fp^{2^\nu}$ to
$\K^K$. Assume $y_k\notin\{0,1\}$ for all $k$, $\lambda,\vartheta\notin\Fp$, $Q<e$, and
\begin{equation}\label{eq:rank}
 \Phi\ \text{is onto }\K^K\quad\text{(as a map of }\Fp\text{-vector spaces)}.
\end{equation}
Then the claimed values $(v_{k,p})_{k,p}$ are uniform in $\K^{Kn_p}$; given them, the mask values
$\mu_k$ are uniform in $\K$; given these, each $u_k$, $k\ge1$, is uniform on the affine subspace
defined by \eqref{eq:zkopen}, $u_0$ is uniform, and the opened columns are uniform subject to the
consistency equations with the revealed combinations. None of these distributions depends on $f$.
\end{lemma}
\begin{proof}[Proof outline]The pads make the evaluations uniform under \eqref{eq:rank}; the masking rows make $\mu_k$ and the revealed combinations uniform; the extension symbols hide the columns (diagonal times Vandermonde). The full proof is in Section~\ref{sup:prf:lem:zkhide} of the supplementary material.\end{proof}

The rank condition \eqref{eq:rank} does not follow from distinct points: it fails for two points
with the same reduced part and different pad coordinates, for Boolean reduced points, for Frobenius
conjugates, and, for every choice of the free coordinates, when $m$ points restricted to a common
Boolean slice with $d$ free coordinates have $2m>2^d$. If some assignment of the free coordinates that
respects the views makes $\Phi$ onto, \eqref{eq:rank} fails with probability at most $2K\nu/p$; we have not
exhibited such assignments for the complete schedule, and \eqref{eq:rank} is a hypothesis of the
zero-knowledge statement (Section~\ref{sup:rank} of the supplementary material).

\phead{Masked Sumchecks}\label{sec:zk-sc}

Every sumcheck is replaced by Libra's zero-knowledge sumcheck~\cite{libra19,cfs17}: to prove
$\sum_xF(x)=H$ the prover holds a mask $G(x)=\sum_ig_i(x_i)$ with random $g_i\in\K[X]$ of degree $7$
(all our combinations have degree at most $6$ per variable), sends $\Gamma=\sum_xG(x)$, receives
$\rho\in\K\setminus\{0\}$, and runs the sumcheck for $\sum_x(F+\rho G)(x)=H+\rho\Gamma$, sending each round
polynomial by its values at $0,\dots,7$; it finally sends $G(q)$. For $\rho=0$ the messages would be
unmasked.

\begin{lemma}[Masked sumcheck~\cite{libra19}]\label{lem:zksc}
If $G(q)$ is correct, the masked sumcheck has soundness error $(7\nu+1)/|\K|$. For $\rho\ne0$ and a
mask $G$ that is uniform and independent of everything else, its messages
$(\Gamma,$ round messages$,G(q))$ can be simulated perfectly given $H$ and $F(q)$.
\end{lemma}

The masks of a phase are slots of one committed \emph{mask pool} $U\in\K^{8S}$, committed as four rows
(components of $U$ and of a uniform $d\in\K^{8S}$ in the basis $1,\sqrt7$). Each $G_s(q_s)$ is a public
linear function $\langle W_s,U\rangle$. At the end of the phase the verifier sends $\lambda$, and with
$W=\sum_s\lambda^sW_s$, $T=\sum_s\lambda^sG_s(q_s)$ the prover sends $\tau=\langle W,d\rangle$, receives
$\rho'\in\K\setminus\{0\}$, and opens $u=U+\rho'd$ with $Q$ columns; the verifier checks
$\langle W,u\rangle=T+\rho'\tau$.

\begin{lemma}[Mask pool]\label{lem:pool}
If some claimed $G_s(q_s)$ is wrong, the verifier accepts with probability at most
$(S_\mathrm{c}+1)/|\K|$ plus the error of the commitment, where $S_\mathrm{c}$ is the number of
masked sumchecks of the phase. In an honest execution, conditioned on $U$, the value $\tau$ is
uniform in $\K$, $u$ is uniform on the affine subspace defined by the checked relation, and the opened
columns are uniform subject to the consistency equations with $u$.
\end{lemma}
\begin{proof}[Proof outline]A wrong value survives the batching with probability $S_{\mathrm c}/|\K|$ and the masked check for one $\rho'$ only; $d$ makes $\tau$ and $u$ uniform. Full proof: Section~\ref{sup:prf:lem:pool} of the supplementary material.\end{proof}

\phead{Every Claim About the Data Is a Zero Sum}\label{sec:zk-zero}

With the pad, a claim about the data half is a claim at pad coordinate $0$, and revealing its value
would reveal the data. All checks are therefore rewritten such that their claimed sum is $0$ and every
value that is revealed is an evaluation at a point whose pad coordinate is a sumcheck challenge.

A zero-check of constraints $C$ on the data runs the masked sumcheck for
\[
 \sum_{(y,x)\in\{0,1\}^{1+\nu}}\eq\bigl((0,r),(y,x)\bigr)\,C\bigl(\hat f(y,x)\bigr)=0 .
\]
Since $\eq((0,r),(y,x))$
vanishes for $y=1$, this is the zero-check of \cref{lem:zerocheck} on the data half, with the same
error; the pad half is unconstrained, and the final evaluations are at $(q_y,q)$ with $q_y$ uniform.
The rounding gadgets, the certifications, the unary lookups and the selector checks of
\cref{lem:max} take this form unchanged.

The product check of \cref{lem:product} reveals the value
$\tilde D_z(r_i,r_j)$. Instead, the parties run one masked sumcheck over $(y,b,l)$ for the zero sum
\begin{equation}\label{eq:zkprod}
\begin{gathered}
 \sum_{y,b,l}\eq\bigl((0,r_b),(y,b)\bigr)\,\Delta(y,b,l)=0,\\
 \Delta=2^{-\nu_k}\tilde D_z(y,b,r_i,r_j)-\mathcal{A}(y,b,l)\,\mathcal{B}(y,b,l),\\
 \mathcal{A}(y,b,l)=\textstyle\sum_i\eq(r_i,i)A_z(y,b,i,l),\\
 \mathcal{B}(y,b,l)=\textstyle\sum_j\eq(r_j,j)B_z(y,b,l,j),
\end{gathered}
\end{equation}
where $b$ indexes batched products and $D_z=\sum_tz^t\hat D^{(t)}$. On the data half \eqref{eq:zkprod} is the
identity of \cref{lem:product} at $(r_b,r_i,r_j)$; the prover reveals the two factors at the final
point, whose reductions to openings of $A$ and $B$ follow at points with uniform pad coordinate.
Row products and the maximum are written in the same way, with the operand factors inside the
sumcheck.

The revealed factors of a product check are not linear in the pads: a factor such as
$\alpha=\sum_i\eq(r_i,i)\tilde A_z(q_y,q_b,i,q_l)$ equals $(1-q_y)\alpha_0+q_y\sum_x w(x)\pi_X(x)$, where
$\alpha_0$ depends only on the data, $\pi_X$ is the pad of the limb table of $A$, and
$w(x)=\eq(\cdot,x)\sum_gz^g\pi_{E_g}(x)$ involves the pads of the bucket indicators. The vector of all
distinct values that the verifier sees outside the sumcheck messages and the openings (the
evaluations of committed tables at all opening points and the factors of all product checks) is
therefore a polynomial function $V(\Pi)$ of degree at most two in the pads $\Pi$. Conditioning on the
indicator pads, as an earlier version of this hypothesis did, would make the indicator evaluations
constant; we state the hypothesis directly about the distribution.

\begin{definition}[Masking hypothesis]\label{def:hyp}
For $\varepsilon_{\mathsf H}\ge0$, hypothesis $\mathsf H(\varepsilon_{\mathsf H})$ holds if for every
witness of $\mathcal R_{\mathrm{model}}$, with probability at least $1-\varepsilon_{\mathsf H}$ over the
verifier coins, the distribution of $V(\Pi)$ over uniform pads is within statistical distance
$\varepsilon_{\mathsf H}$ of the uniform distribution on $\K^{N_{\mathrm v}}$, where $N_{\mathrm v}$ is the
number of distinct revealed values.
\end{definition}

A sufficient condition is a \emph{fresh-pad order} of the revealed values, in which each value is an
affine function, onto $\K$, of pad entries that occur in no earlier value (Section~\ref{sup:hyp} of the
supplementary material). We have not constructed one for the complete schedule and do not bound
$\varepsilon_{\mathsf H}$; proving $\mathsf H$, or replacing it by independent masks per revealed value, is the
main open step of the security analysis.

\phead{Lookups Without Revealing Helper Sums}\label{sec:zk-lookup}

Instead of revealing multiplicities and helper sums, each phase commits to the multiplicity table
$m$ (before $\gamma,\xi$), then to all helpers and the table-side helper $h_T=m/(\xi-t)$, and runs two
masked zero sums: the zero-check of $h_T(\xi-t)-m$ on the table domain, and one sumcheck of
$\sum_{y,x}(1-y)\bigl[\sum_\ell\iota_\ell h_\ell(y,x)-\iota_Th_T(y,x)\bigr]=0$ over the union of all
domains ($\iota_\ell$ embeds a domain into the low variables). On the data half they state
\cref{lem:logup} at $\xi$; only masked messages and evaluations at points with uniform pad coordinate
are revealed.

\phead{The Complete Model}\label{sec:zk-model}

The prompt is public, thus the gather $e_i=\mathrm{wte}_{\tau_i}$ is a sum of
views with fixed row indices. With a challenge $\kappa$ and $P=X+\sum_g\kappa^{g+1}E_g$ (all committed
tables of an encoding), the parties run the masked zero sum
\[
 \sum_{y,i,d}\eq\bigl((0,r),(y,i,d)\bigr)\bigl[P_e(y,i,d)-P_{\mathrm{wte}}(y,\tau_i,d)\bigr]=0 .
\]
At the final point the second term needs one evaluation of the vocabulary commitment per distinct
token; on the data half the zero sum states that $e$ equals the selected rows of $\mathrm{wte}$, and
$x_0=\RN(e+\mathrm{wpe}_{0:T})$ is an element-wise sum through a slice view.

After the last block and the final LayerNorm $h_f$, the
vocabulary is split into chunks $\mathrm{wte}_c$ of $2^{13}$ rows, and
$\ell_c=\RN(h_f\,\mathrm{wte}_c^{\mathsf T})$ is a matrix product whose second operand is a transpose
view of the chunk. The last row of every $\ell_c$ is compared with the public logits by the masked
zero sum $\sum\eq((0,r),(y,j))\bigl[P_{\ell_c}(y,T-1,j)-P_{y}(j)\bigr]=0$, in which the verifier
evaluates the public side itself.

The argument runs in phases that share one transcript (embedding, one per block, final LayerNorm,
one per vocabulary chunk). Every tensor is committed once, in the phase that produces it, and
opened once, at the end of the last phase that uses it, at all points of all its claims; each phase
has its own lookup challenges, multiplicities and mask pool, and all sizes are fixed by the plan.

\begin{theorem}\label{thm:zk}
Let $\Pi_{\mathrm{model}}$ be the interactive argument for $\mathcal{R}_{\mathrm{model}}$ described
above, with the hash function modelled as a random oracle.
\begin{enumerate}[(a),leftmargin=*,nosep]
\item $\Pi_{\mathrm{model}}$ is complete, provided \eqref{eq:inv} holds for every gadget and every
 lookup and table has fewer than $p$ entries.
\item Its soundness error is at most the sum, over all phases, of the errors of \cref{thm:block} for
 the steps of the phase (with \cref{lem:zksc} in place of \cref{lem:sumcheck} and the zero-sum forms
 above), of the two lookup sumchecks and \cref{lem:logup}, of \cref{lem:pool}, of the gather and
 output zero sums, and of \cref{lem:zkbind} for every opening.
\item Assume $\mathsf H(\varepsilon_{\mathsf H})$ (\cref{def:hyp}) and the leaf-entropy condition
 $\mathsf E(\varepsilon_{\mathrm{RO}})$ stated after the proof. Then $\Pi_{\mathrm{model}}$ is statistical
 honest-verifier zero knowledge with error
 \[
 \varepsilon_{\mathrm{zk}}\le\varepsilon_{\mathsf H}+\frac{2\nu\sum K}{p}
 +\frac{N_{y}}{p}+\varepsilon_{\mathrm{RO}},
 \]
 where $\sum K$ is the total number of opening points over all commitments, $N_y$ the number of
 opening points, and $\varepsilon_{\mathrm{RO}}$ is the oracle-simulation error of $\mathsf E$.
\end{enumerate}
\end{theorem}
\begin{proof}[Proof outline](a) and (b) as for \cref{thm:block}, with the masked components; (c) a simulator samples the revealed values under $\mathsf H$, simulates sumchecks by \cref{lem:zksc}, openings by \cref{lem:zkhide,lem:pool}, and programs the oracle under $\mathsf E$. The full proof is in Section~\ref{sup:prf:thm:zk} of the supplementary material.\end{proof}

The leaf-entropy condition $\mathsf E(\varepsilon_{\mathrm{RO}})$ asks that every leaf message or
internal-node input that the simulator programs have conditional guessing probability at most
$\varepsilon_{\mathrm{RO}}/(q_HN_{\mathrm u})$ given the complete view before it is programmed, for $q_H$
oracle queries and $N_{\mathrm u}$ programmed inputs; we do not discharge it for the complete schedule.
With $\sum K\approx2.2\cdot10^3$ and $\nu\le25$ the linear-algebra terms of (c) evaluate to about
$2^{-47.2}$; $\varepsilon_{\mathsf H}$ and $\varepsilon_{\mathrm{RO}}$ remain symbolic, and we claim no
end-to-end privacy level. The interactive soundness error of the complete model is below $2^{-85}$,
dominated by the proximity test of the openings ($(1-\delta/3)^{250}<2^{-101}$ per opening) and by
lookup compression; we give no bound for the Fiat--Shamir implementation. The formats reveal a
magnitude bound per tensor and must be fixed by calibration in deployment, and binding many proofs to
one published parameter commitment needs $e\ge NQ+1$ and the rank condition across proofs; these
points are detailed in Section~\ref{sup:zkmore} of the supplementary material.

%% file: eval.tex
\section{Evaluation}\label{sec:eval}

We use GPT-2 small~\cite{gpt2} with its released weights and WikiText-2~\cite{wikitext}. A NumPy
forward pass runs under four semantics: \textsf{fp32}; \textsf{native} (\bfs{}, binary32 accumulation
in BLAS order); \textsf{native$'$} (reversed order); and \textsf{CR} (correctly rounded products by the
Ziv procedure); element-wise layers are evaluated in binary64 and rounded once, so these experiments
measure correctly rounded products, while the block and complete-model runs below measure the exact relation.
\textsf{CR-w} uses calibrated windowed formats $\Fl(e_0,48)$, $w=12$, four buckets. The protocol is
implemented as a Python reference over $\Fp$ ($w=12$, $n_c=4$, satisfying \eqref{eq:pc} for $k\le8192$)
that exercises every constraint, and as a Rust prover (quadratic extension, rate-$1/4$ codes, BLAKE3,
250 queries). All runs use two virtual cores and 7~GB of memory.

\phead{Native Kernels versus the Canonical Semantics}

\begin{table}[!htb]
\centering\footnotesize\setlength{\tabcolsep}{2.5pt}
\caption{GPT-2 matrix products (4 sequences of 128 tokens, 12 layers): fraction of outputs where
the native kernel differs from CR, and where the Ziv test with the bound of \cref{prop:higham} fails.}\label{tab:e1}
\begin{tabular}{lrrrrr}
\toprule
 & & & native & \multicolumn{2}{c}{Ziv test fails}\\
\cmidrule(lr){5-6}
site & $k$ & entries & $\ne$ CR & bin32 & bin64\\
\midrule
\textsf{qkv}   & 768  & 14{,}155{,}776 & $1.4\cdot10^{-4}$ & $45.3\%$ & $6.4\cdot10^{-7}$\\
\textsf{qk}    & 64   &  9{,}437{,}184 & $3.1\cdot10^{-5}$ & $1.5\%$  & $1.0\cdot10^{-5}$\\
\textsf{av}    & 128  &  4{,}718{,}592 & $3.5\cdot10^{-5}$ & $2.0\%$  & $1.7\cdot10^{-4}$\\
\textsf{aproj} & 768  &  4{,}718{,}592 & $3.0\cdot10^{-4}$ & $60.5\%$ & $0$\\
\textsf{fc}    & 768  & 18{,}874{,}368 & $7.5\cdot10^{-5}$ & $32.5\%$ & $5.3\cdot10^{-7}$\\
\textsf{mproj} & 3072 &  4{,}718{,}592 & $2.9\cdot10^{-4}$ & $99.7\%$ & $4.2\cdot10^{-7}$\\
\textsf{logits}& 768  & 25{,}731{,}584 & $4.8\cdot10^{-5}$ & $2.6\%$  & $0$\\
\midrule
all            &      & 82{,}354{,}688 & $9.5\cdot10^{-5}$ & & \\
\bottomrule
\end{tabular}
\end{table}

\Cref{tab:e1} reports, for every class of matrix product in the model, how often the native kernel
differs from the correctly rounded result. Over $8.2\cdot10^7$ entries the rate is
$9.5\cdot10^{-5}$, about one entry in ten thousand, and at most $3\cdot10^{-4}$ for any class. We
did not record the distance where they differ, which \cref{prop:determined} does not bound.

The last two columns measure the Ziv rounding test. With binary64 accumulation it fails on at most
$1.7\cdot10^{-4}$ of the entries, thus the canonical result can be computed from one binary64 product,
one product of absolute values and a negligible number of exact recomputations. With binary32
accumulation the worst-case bound of \cref{prop:higham} flags $1.5\%$--$99.7\%$ of the entries; a
certificate tight for binary32 accumulation is open.

Two native summation orders disagree with each other as often as either disagrees with the
correctly rounded result (Table~\ref{tab:e2} of the supplementary material): the canonical semantics is
one more member of the family of native kernels, defined without reference to an order.

\begin{figure}[t]
\centering
\includegraphics[width=\columnwidth]{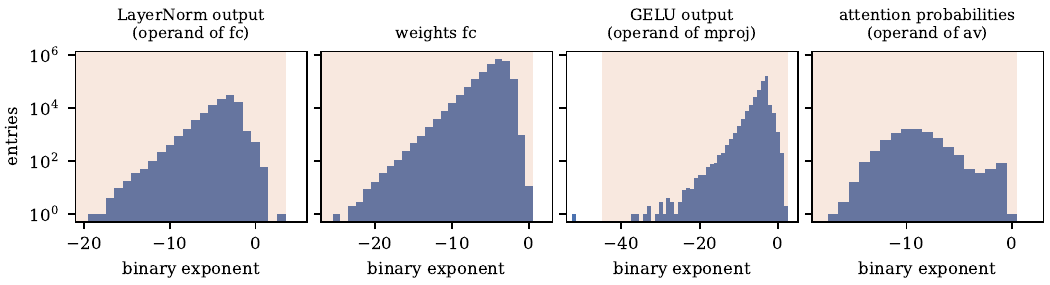}
\caption{Binary exponents of the non-zero entries of four GPT-2 operands in layer 5 (128 tokens,
logarithmic vertical axis). The shaded band is a window of 48 bits below the largest exponent.
Weights and LayerNorm outputs fit; GELU outputs and attention probabilities have thin tails of very
small values.}
\label{fig:exp}
\end{figure}

\emph{Exponent ranges.} Across all operands and results of all matrix products, the span of binary
exponents plus the eight significand bits is 31--45 bits, except GELU outputs (65 bits) and
attention probabilities (141 bits), whose thin tails of tiny values (\cref{fig:exp}) leave a 48-bit window for
$7\cdot10^{-5}$ and $1.8\%$ of their entries. Windowed formats with $G=4$ everywhere change only these
tail entries, and the final logits are bit-identical to those of unwindowed \bfs{} at all $4{,}096$
positions (Section~\ref{sup:exp} of the supplementary material).

\phead{Model-Level Behaviour}\label{sec:model}

\begin{table}[!htb]
\centering\footnotesize\setlength{\tabcolsep}{2.5pt}
\caption{GPT-2 on WikiText-2: 16 sequences of 256 tokens, $4{,}080$ scored tokens. Log-loss in
nats; the standard error of the mean log-loss itself is $0.05$. For comparison, \textsf{native}
and \textsf{native$'$} choose the same greedy token at $91.9\%$ of the positions.}\label{tab:e3}
\begin{tabular}{lrrrr}
\toprule
 & perp- & $\Delta$ log-loss vs. & greedy & logits\\
variant & lexity & \textsf{fp32} (mean$\pm$s.e.) & $=$\textsf{CR} & $=$\textsf{CR}\\
\midrule
\textsf{fp32}            & $45.99$ & ---                    & $88.7\%$ & ---\\
\textsf{native}          & $46.05$ & $+0.0013\pm0.0031$     & $92.1\%$ & $53.2\%$\\
\textsf{native$'$}       & $46.02$ & $+0.0006\pm0.0031$     & $92.2\%$ & $52.3\%$\\
\textsf{CR}              & $46.14$ & $+0.0032\pm0.0032$     & $100\%$  & $100\%$\\
\textsf{CR-w} (48-bit)   & $46.14$ & $+0.0032\pm0.0032$     & $100\%$  & $100\%$\\
\bottomrule
\end{tabular}
\end{table}

\Cref{tab:e3}: the four semantics are statistically indistinguishable in perplexity (paired
differences at most $0.0032\pm0.0032$ nats). The greedy token is far more sensitive: two native orders
agree at $91.9\%$ of the positions and \textsf{CR} agrees with each at $92\%$, because one-ulp differences
are amplified and \bfs{} logits near $100$ have spacing $0.5$; \textsf{CR} is no further from either
native kernel than they are from each other.

\phead{What an Error-Bounded Relation Would Give Away}\label{sec:slack}
An error-bounded relation that accepts unmodified native kernels lets every output be any
$\RN(x)$ within the certified bound of \cref{prop:higham}. For binary32 accumulation this leaves a
choice at $41\%$ of the product outputs, and a first-order adversary that coordinates these
one-ulp choices across all products changes the greedy next token at 47 of 48 targets (median logit
gain $3.5$); random choices within the same slack flip one target, and with a binary64-level bound
the adversary gains nothing. Such a relation therefore does not support claims about generated
tokens, whereas the canonical relation has no slack (details in Section~\ref{sup:slack} of the
supplementary material).

\emph{The protocol on real operands.} The reference verifier accepts seven instances built from
operands captured during a \textsf{CR-w} forward pass ($k\le3072$; 52--62 committed cells and 32--42
lookups per output), 25 further honest instances, and all 28 matrix products of layer~5 at full width
($1.14\cdot10^8$ multiply--adds, $7.4\cdot10^6$ committed cells against $7.3\cdot10^9$ constraints at 64 per
multiply--add). It rejects all 18 hand-built and 239 randomized false statements and malformed
witnesses, including wrong neighbours at ties, binade boundaries, cancellation and underflow, and
re-balanced limb witnesses (Section~\ref{sup:piop} of the supplementary material).

\phead{A Native Prover: Every Matrix Product of GPT-2}\label{sec:native}

\begin{table}[!htb]
\centering\footnotesize\setlength{\tabcolsep}{2.5pt}
\caption{Native prover, two cores. ``Block'' is layer~5; ``model'' is all blocks and the unembedding.
Setup (once per model) commits to and certifies the parameters.}\label{tab:native}
\begin{tabular}{lrrrrrr}
\toprule
 & inst- & & prover & verifier & proof & setup\\
workload & ances & MACs & (s) & (s) & (MB) & (s)\\
\midrule
block, 4 tokens   & 28  & $2.8\cdot10^{7}$ & $16.8$ & $0.44$ & $20$  & $42$\\
block, 16 tokens  & 28  & $1.14\cdot10^{8}$ & $23.4$ & $0.72$ & $34$  & $40$\\
block, 64 tokens  & 28  & $4.6\cdot10^{8}$ & $52.6$ & $1.34$ & $54$  & $41$\\
model, 16 tokens & 343 & $1.98\cdot10^{9}$ & $371.6$ & $9.5$ & $449$ & $625$\\
\bottomrule
\end{tabular}
\end{table}

For a 16-token input all 343 products of GPT-2 small ($1.98\cdot10^9$ multiply--adds) are proved in
$372$ seconds and verified in $9.5$ seconds with a $449$~MB proof (\cref{tab:native}); each additional
token costs $0.61$ seconds per block. Proofs are large because every table has its own $O(\sqrt N)$
commitment. Wrong outputs, altered witness cells and altered proof elements are rejected.

\phead{An End-to-End Transformer Block}\label{sec:blockrun}

We proved the first block of GPT-2 small on the 16-token input ``The capital of France is Paris.
The capital of Germany is Berlin, and the'' with the steps of \cref{tab:steps}. Each of the 41
tensors (14 parameter tensors, the input, 25 intermediate and output tensors and the selector of
\cref{lem:max}) has a public format $\Fl(e_0,48)$ with four 12-bit buckets, $e_0$ chosen from this
input with a margin of two bits; a deployment would fix the formats by calibration.

The whole block, including LayerNorm, the causal softmax and the chaining of all tensors, is proved
in 42.9 seconds and verified in 0.96 seconds with a 43.2~MB proof ($1.34\cdot10^8$ committed field elements, 143 evaluation claims, 5.9~GB peak memory); every
commitment is opened once, in a batch, at all 143 points at which it is used. Relative to binary64,
the error of the block output is $0.43\%$ for the canonical block, against $0.37\%$ for correctly
rounded and for native \bfs{} inference, all at the scale of the unit roundoff $2^{-8}$; the canonical
form rounds five times inside LayerNorm where the reference rounds once.
The verifier rejects all 55 attacks on the block argument (Table~\ref{tab:tests} of the
supplementary material): 26 single changed values, 19 executions in which one layer is wrong and every
later layer is recomputed honestly from it, which is the attack that chaining must stop, and 10
altered proof elements.

\input{eval_zk}

\begin{figure}[t]
\centering
\includegraphics[width=0.85\columnwidth]{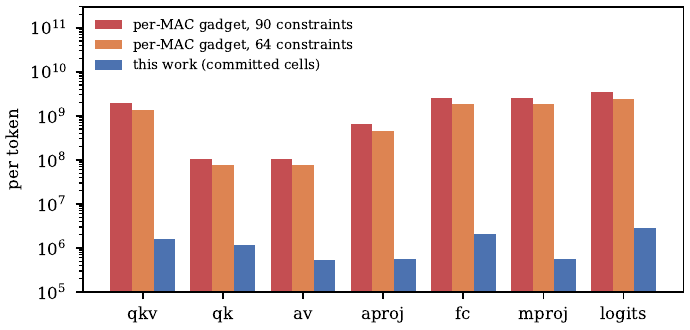}
\caption{Committed cells of $\Pi_{\CR}$ and constraint counts of per-operation floating-point
gadgets, per generated token of GPT-2 small at context length 128, by class of matrix product
(logarithmic axis).}
\label{fig:cost}
\end{figure}

\phead{Cost and Comparison}\label{sec:cost}
Over all matrix products of one GPT-2 forward step (context length 128) the protocol commits to
$9.4\cdot10^{6}$ cells per token, against $8.1\cdot10^{9}$ and $1.1\cdot10^{10}$ constraints for
per-operation gadgets at 64 and 90 constraints per multiply--add, a ratio of 861 to 1211 (\cref{fig:cost}). The only
system that proves floating-point linear layers, ZIP~\cite{zip25}, reports $34.5$ hours on 48 cores
for an 11M-parameter model, against our $372$ seconds on two cores for all products of a
124M-parameter model; integer-model provers~\cite{deepprove26} are within a small constant factor of
our marginal cost per token. These comparisons are in different units and on different hardware;
the cost model, the published figures and a comparison of approaches are in
Section~\ref{sup:cost} of the supplementary material.

\section{Related Work}\label{sec:related}

Proofs of floating-point computations treat one operation at a time: Garg et al.~\cite{garg22},
conversions between representations~\cite{mystique21}, IEEE-compliant circuits at 64 constraints per
multiplication~\cite{ernstberger24}, and ZIP~\cite{zip25}, whose linear layers reuse these circuits.
We remove the dependence on the number of operations inside a matrix product by proving its
correctly rounded value. Integer-model systems~\cite{peng25,zkllm24,zkgpt25,deepprove26,openllm26,joltatlas26,nanozk26,flashprove26,sash26}
prove a stand-in model; sampling~\cite{anchuri26,peigne26} and trusted hardware~\cite{veriattn26}
trade soundness or trust for speed, and bit-exact re-execution~\cite{cankaya26} shows how much
device-specific detail a native relation must fix. Correctly rounded and reproducible dot
products~\cite{ogita05,demmel13,muller18}, Ziv's test~\cite{ziv91}, the error analysis of~\cite{higham02}
and Freivalds-type checks~\cite{freivalds77,ji17} are classical; their combination with sumcheck-based
arguments is, to our knowledge, new. We use the sumcheck of~\cite{thaler13}, logarithmic-derivative
lookups~\cite{haboeck22,papini23,lasso} and hash-based commitments~\cite{fri18,basefold24,whir25,binius23}.

\section{Conclusion}\label{sec:conclusion}

Proving the correctly rounded matrix product, rather than the rounding sequence of a kernel, turns
floating-point matrix multiplication into an integer identity plus one rounding per output, at the
asymptotic cost of integer-model provers up to the number of exponent buckets, with a semantics that
is canonical and empirically indistinguishable from native \bfs{} inference. Open problems are a
proof of the masking hypothesis for the complete transcript, a polylogarithmic commitment (the
proof of the complete model is 0.83~GB), a certificate tight for binary32 accumulation, certified
unary tables, and binding many proofs to one published parameter commitment.

%% file: eval_zk.tex
\phead{The Complete Model with the Privacy Construction}\label{sec:zkrun}

We proved the complete forward pass of GPT-2 small (12 blocks, embeddings, final LayerNorm and the
logits over the vocabulary of 50{,}257 tokens, padded to $7\cdot2^{13}$) on the same prompt with the
argument of \cref{sec:zk}. The statement is the prompt, the plan with 482 tensor formats and 61 unary
tables, and the 57{,}344 padded logits of the last position; the 124 million parameters and all other
tensors are hidden. The commitment uses codes of rate below $0.266$, $Q=250$ queries and $e=256$
extension symbols. Encodings are never stored: leaves are hashed while rows are encoded in batches,
queried columns are recomputed at opening, and the vocabulary chunks are committed in the embedding
phase and recomputed in their logits phase.

\begin{table}[!htb]
\centering\footnotesize\setlength{\tabcolsep}{3pt}
\caption{Complete GPT-2 forward pass with the privacy construction, 16 tokens, two cores; block and
chunk rows are means.}\label{tab:zkmodel}
\begin{tabular}{lrrrr}
\toprule
 & prover & proof & mask & stream\\
phase & (s) & (MB) & slots & elements\\
\midrule
embedding & 70.2 & 5.2 & 88 & 868\\
one block & 67.4 & 59.9 & 930 & 11561\\
final LayerNorm & 2.8 & 6.9 & 152 & 2082\\
one vocabulary chunk & 33.9 & 16.5 & 146 & 1472\\
\midrule
complete model (21 phases) & 1121 & 846 & 12422 & 151988\\
\bottomrule
\end{tabular}
\end{table}

The complete model is proved in 18.7 minutes and verified in 21 seconds with a 0.83~GB proof and
4.4~GB peak memory (\cref{tab:zkmodel}); a block costs 67.4 seconds against $42.9$ without the privacy
construction, mostly because the pads double the committed data. The logits differ from binary64 by
0.37\% (native \bfs{}: 0.52\%), and at the public last position all semantics predict the same
token. The verifier rejects all 236 attacks: 25 single changed computed tensors, 8 consistent
executions recomputed from one wrong value including the public logits, and 203 tampered proofs
(Table~\ref{tab:tests} and Section~\ref{sup:attacks} of the supplementary material).

%% file: supp_proofs.tex
\subsection{Proof of \cref{lem:round}}\label{sup:prf:lem:round}
\begin{proof}
First let $c=0$, hence $\sigma=1$, $M=0$, $Q=0$, $u=2^{e_0}$, $\lambda=u/2$ and $\rho=v$. The
neighbours of $0$ are $\pm u$, both with odd significand $1$, thus $\RN(v)=0$ exactly when
$|v|\le u/2$, which is (a); (b) is vacuous.

Now let $c\ne0$. Since $\Fl(e_0,H)$ and the tie rule are symmetric under negation, we may assume
$\sigma=+1$, hence $c>0$ and $\rho=v-c$. Let $c^-<c$ be the predecessor of $c$ in $\Fl(e_0,H)$. If
$M>128$, or $Q=0$ (then $M\ge1$), the predecessor is $(M-1)2^{Q+e_0}=c-u$. If $M=128$ and $Q\ge1$,
it is $255\cdot2^{Q-1+e_0}=c-u/2$. In both cases $(c^-+c)/2=c-\lambda$. Let $c^+>c$ be the
successor, if it exists; it is $c+u$ (for $M=255$ this is $128\cdot 2^{Q+1+e_0}$), thus
$(c+c^+)/2=c+u/2$. If $c$ is the largest element, then $M=255$, $Q=H-1$ and $c+u/2=\Omega$.

By definition of rounding to nearest, $\RN(v)=c$ if and only if $v$ lies in the closed interval
$[c-\lambda,\,c+u/2]$, where an endpoint that is a midpoint between $c$ and a neighbour is included
exactly when $c$ wins the tie, and the endpoint $\Omega$ (when $c$ is the largest element) is
excluded. Neighbours have significands of opposite parity, hence $c$ wins a tie exactly when $M$ is
even; and when $c$ is the largest element $M=255$ is odd, thus the exclusion of $\Omega$ is also
expressed by (b). Hence $\RN(v)=c$ with $|v|<\Omega$ if and only if (a) and (b) hold.
\end{proof}

\subsection{Proof of \cref{prop:orders}}\label{sup:prf:prop:orders}

\begin{proof}
Let $A=(2^{24},2^{16},1,1)$ and $B=(1,1,1,1)^{\mathsf T}$; all entries are \bfs{} numbers. The exact
value is $v=2^{24}+2^{16}+2$. In the binade $[2^{24},2^{25})$ the spacing of binary32 is $2$ and
that of \bfs{} is $2^{17}$, thus the \bfs{} breakpoint between $2^{24}$ and $2^{24}+2^{17}$ is
$2^{24}+2^{16}$, and $\RN(v)=2^{24}+2^{17}$. Left-to-right accumulation computes $2^{24}+2^{16}$
exactly; adding $1$ is a tie in binary32 between $2^{24}+2^{16}$ and $2^{24}+2^{16}+2$, which goes
to the even significand $2^{24}+2^{16}$, twice. The accumulated value $2^{24}+2^{16}$ is exactly
the \bfs{} breakpoint, a tie that goes to the even significand, hence the output is $2^{24}$.
Right-to-left accumulation computes $1+1=2$, then $2+2^{16}$ and $2+2^{16}+2^{24}$ exactly, and
outputs $2^{24}+2^{17}$.
\end{proof}

\subsection{Proof of \cref{lem:product}}\label{sup:prf:lem:product}
\begin{proof}
On the cube, $A_z(i,l)=X^A_{il}z^{g^A_{il}}$ because $E^A$ is one-hot, hence by \cref{lem:bucket},
reduced modulo $p$, the matrix $A_zB_z$ over $\K$ equals $\sum_tz^tD^{(t)}$. Suppose
$\hat D^{(t)}_{ij}\ne D^{(t)}_{ij}$ in $\Fp$ for some $(t,i,j)$. Then
$\sum_tz^t(\hat D^{(t)}_{ij}-D^{(t)}_{ij})$ is a non-zero polynomial in $z$ of degree at most
$T_D-1$, thus except with probability $(T_D-1)/|\K|$ the matrices $\sum_tz^t\hat D^{(t)}$ and
$A_zB_z$ differ; then their multilinear extensions differ at $(\rho_I,\rho_J)$ except with
probability $(\nu_m+\nu_n)/|\K|$. Since
$\widetilde{A_zB_z}(\rho_I,\rho_J)=\sum_l\tilde A_z(\rho_I,l)\tilde B_z(l,\rho_J)$~\cite{thaler13},
the claim~\eqref{eq:thaler} is then false. Its summand has degree $2$ in each of $\nu_k$ variables,
hence by \cref{lem:sumcheck} the verifier rejects or the final value differs from
$\tilde A_z(\rho_I,\rho_L)\tilde B_z(\rho_L,\rho_J)$, except with probability $2\nu_k/|\K|$. In the
latter case $a$ or $b'$ is not the true value. The true value is
$\tilde A_z(\rho_I,\rho_L)=\sum_x\eq((\rho_I,\rho_L),x)X^A(x)\sum_gz^gE^A_g(x)$, a sum over
$\nu_m+\nu_k$ variables of a polynomial of individual degree $3$, thus a false claim for $a$ is
accepted with probability at most $3(\nu_m+\nu_k)/|\K|$, and likewise for $b'$. The union bound
gives $\varepsilon_{\mathrm{prod}}$. Finally $|D^{(t)}_{ij}|\le k\Xmax^2<p/2$, hence equality in $\Fp$
identifies the integer.
\end{proof}

\subsection{Proof of \cref{lem:gadget}}\label{sup:prf:lem:gadget}
\begin{proof}
\emph{Soundness.} \emph{Decoding.} By (L1) there are $\sigma$, $M$, $r$ with $X=\sigma M2^r$,
$P=2^r$, $\sg=\sigma$ and the stated meaning of $\ev,\bd,\sub,\rz$, and either $M\ge128$, or $M<128$
and $r=0$. By (C1) each $E_g$ is a bit and, since $G_C<p$, exactly one $E_g$ is $1$; let $g$ be its index and $Q=wg+r\in[H_C]$. If $M<128$
then $\sub=1$ and (C2) forces $g=0$, hence $Q=0$. Hence $(\sigma,M,Q)$ is canonical and encodes
$c=X\beta^{\tau}2^{e_A+e_B}\in\Fl_C$, with $\tau=g+d$ and $U=P\beta^\tau$.

\emph{The residual.} By (L2)--(L4) and (C3) the columns $\tp,a,\lt,b',b_t$ are integers with
$\tp\in\{0,1\}$, $a\in[4P]$, $\lt=[a<P]$, $b'\in[\beta/4]$, $b_t\in[\beta]$. The sum in (C4) is the
indicator of $\tau-1\le t$, thus $b_t=0$ for $t\ge\tau-1$; and by (C3), $\tp=1$ forces $a=b'=b_t=0$.
Define $V$ by \eqref{eq:V}. If $\tp=1$ then $V=U$. If $\tp=0$,
\[
 0\le V\le(4P-1)\frac{\beta^\tau}{4}+\Bigl(\frac{\beta}{4}-1\Bigr)\beta^{\tau-1}+\bigl(\beta^{\tau-1}-1\bigr)=U-1 .
\]
Hence $0\le V\le U$, with $V=U$ if and only if $\tp=1$. Furthermore, for $\tp=0$ the part of $V$
below $a\beta^\tau/4$ is less than $\beta^\tau/4$, hence $V\ge U/4=P\beta^\tau/4$ if and only if
$a\ge P$; and $V=0$ if and only if $\tp=0$ and $S=0$.

\emph{The carry chain.} In (C7), $c_{t}$ is an integer in $[-\beta^{n_c}/2,\beta^{n_c}/2)$ by (L4),
and $e_t$ is the residue of an integer of absolute value at most $\Emax$: $D^{(t)}$ is a true limb,
$|D^{(t)}|\le k\Xmax^2$ by \cref{lem:bucket}; $|X^b|,|X|\le\Xmax$; $P\le\beta/2$;
$a\beta/4<\beta^2/2$; $b'<\beta/4$; $b_t<\beta$; and the indicators are bits. Hence
$e_t+c_{t-1}-\beta c_t$ is an integer of absolute value at most
$\Emax+(\beta+1)\beta^{n_c}/2<p$ that is divisible by $p$; it is zero. Multiplying the $t$-th
equation by $\beta^t$ and summing over $t\in[T_c]$, the carries telescope:
\[
 \sum_{t\in[T_c]}e_t\beta^t=\sum_{t\in[T_c]}(\beta c_t-c_{t-1})\beta^t=c_{T_c-1}\beta^{T_c}-c_{-1}=0 .
\]
The indicator $E_{t-d}$ selects $t=\tau$, $E_{t-d+1}$ selects $t=\tau-1$, and $E^b_{t-d_b}$ selects
$t=g^b+d_b$; all three positions and all limbs of $D$ lie in $[T_c]$ by the definition of $T_c$.
With \eqref{eq:N} and \eqref{eq:V} the left side is therefore
\[
 2N-(2X-\sigma P)\beta^\tau-2\sigma V ,
\]
hence $2N=(2\sigma M-\sigma)U+2\sigma V$, that is, $\sigma N=(M-\tfrac12)U+V$.

\emph{Rounding.} With $v=N2^{e_A+e_B}$ we get $\rho=\sigma(v-c)=(V-U/2)2^{e_A+e_B}$ and
$u=U2^{e_A+e_B}$. From $0\le V\le U$, $-u/2\le\rho\le u/2$. If $M=128$ and $Q\ge1$, then $\bd=1$ and
$E_0\rz=0$, thus (C5) gives $\tp=1$ or $\lt=0$, hence $V\ge U/4$ and $\rho\ge-u/4$. This is (a) of
\cref{lem:round}. If $M$ is odd, $\ev=0$ and (C6) gives $\tp=0$, hence $V\ne U$, and
$S\cdot\inv=1$ in $\Fp$, thus $S\not\equiv0\pmod p$, hence the integer $S$ is non-zero and $V\ne0$. An odd $M$ is not $128$, hence $\lambda=u/2$ and both inequalities of (a) are strict, which
is (b). By \cref{lem:round}, $|v|<\Omega_C$ and $c=\RN(v)$.

\emph{Completeness.} Let $(\sigma,M,Q)$ be the canonical triple of $c=\RN(v)$ and set the encoding
columns accordingly; (L1), (C1), (C2) hold. Let $V=\sigma N-(M-\tfrac12)U$, an integer. By
\cref{lem:round}, \eqref{eq:Vcond} holds. If $V=U$ set $\tp=1$ and $a=b'=b_t=0$. Otherwise set
$\tp=0$, $a=\lfloor V/(\beta^\tau/4)\rfloor<4P$, let $b'$ be the quotient of the remainder by
$\beta^{\tau-1}$, which is less than $\beta/4$, and let $b_t$, $t<\tau-1$, be the base-$\beta$
digits of what is left; set $b_t=0$ for $t\ge\tau-1$ and $\lt=[a<P]$. Then \eqref{eq:V}, (L2)--(L4),
(C3), (C4) hold. If $M=128$ and $Q\ge1$, then $V\ge U/4$, thus $\tp=1$ or $a\ge P$, which is (C5); in
all other cases $\bd(1-E_0\rz)=0$. If $M$ is odd, $V\notin\{0,U\}$, hence $\tp=0$ and $S\ge1$; since
$S<4P+\beta/4+T_b\beta\le(T_b+3)\beta<p$ by \eqref{eq:inv}, $S$ is invertible in $\Fp$; set
$\inv=S^{-1}$. If $M$ is even set $\inv=0$. This is (C6).
By construction $\sum_{t\in[T_c]}e_t\beta^t=2N-(2X-\sigma P)\beta^\tau-2\sigma V=0$. Define
$c_t=\beta^{-(t+1)}\sum_{s\le t}e_s\beta^s$. This is an integer because
$\sum_{s\le t}e_s\beta^s=-\sum_{s>t}e_s\beta^s$ is divisible by $\beta^{t+1}$; it satisfies
$e_t+c_{t-1}-\beta c_t=0$ and $c_{T_c-1}=0$; and
$|c_t|\le\Emax\sum_{s\le t}\beta^{s-t-1}<\Emax/(\beta-1)<\beta^{n_c}/2$, thus its shifted digits exist.
This is (C7).
\end{proof}

\subsection{Proof of \cref{thm:main}}\label{sup:prf:thm:main}
\begin{proof}
\emph{Completeness.} The honest prover computes the limbs $D^{(t)}$ of \eqref{eq:D}, which satisfy
\eqref{eq:thaler} for every $z$ by \cref{lem:bucket}, hence the product check passes. For every entry
it computes the assignment of \cref{lem:gadget}(2), which satisfies all constraints and lookups, thus
the gadget check passes. Canonical operand encodings pass the operand check.

\emph{Soundness.} Suppose the verifier accepts with probability greater than $\varepsilon$. Except
with probability $\varepsilon_{\mathrm{op}}$ the operand oracles have one-hot bucket indicators and
limbs in $\mathcal{L}$, hence they encode matrices $A\in\Fl_A^{m\times k}$, $B\in\Fl_B^{k\times n}$ and
$b\in\Fl_b^{n}$; one-hotness of the $G_\Theta$ indicator columns and the lookup helper constraint form a zero-check
of $G_\Theta+2$ constraints of degree $2$, whence the bound by \cref{lem:zerocheck}. By
\cref{lem:product}, except with probability $\varepsilon_{\mathrm{prod}}$ the oracles
$\hat D^{(t)}$ hold the residues of the true limbs. By \cref{lem:zerocheck} with $\delta=5$, except
with probability $(q-1+7(\nu_m+\nu_n))/|\K|$ every output row satisfies (C1)--(C7) over $\Fp$, and
except with probability $\varepsilon_{\mathrm{lk}}$ it satisfies (L1)--(L4). Then
\cref{lem:gadget}(1) applies to every entry: the output oracles are the canonical encoding of a
matrix $C$ with $C_{ij}=\RN(N_{ij}2^{e_A+e_B})$, and $N_{ij}2^{e_A+e_B}=v_{ij}$ by
\cref{lem:bucket}. Hence the statement is in $\mathcal{R}_{\CR}$, a contradiction.

\emph{Efficiency.} The count \eqref{eq:cells} lists the encoding of $C$ (limb, $G_C$ indicators and
six attributes), the columns $\tp,a,\lt,b',\inv$, the $T_b$ digits, the $T_D$ limbs and the carry
digits. The prover's field work is the sumchecks of \cref{lem:product}, linear in the sizes of
their tables, and the zero-check, linear in the size of the output table for constant degree.
\end{proof}

\subsection{Proof of \cref{lem:max}}\label{sup:prf:lem:max}
\begin{proof}
By (i) each $\Sigma_{ij}$ is a bit and vanishes above the diagonal. Condition (ii) is a polynomial
identity of degree $1$ in $\lambda$; unless it holds coefficientwise, a random $\lambda$ satisfies it with
probability at most $1/|\K|$. Coefficientwise it says $\sum_j\Sigma_{ij}=1$ in $\Fp$; the left side is an
integer between $0$ and the row length, which is less than $p$ by (iv), hence it equals $1$ and exactly
one $j^*\le i$ is selected (without (iv), $p+1$ selected entries would also sum to $1$), and $X^S_{ij^*}z^{g^S_{ij^*}}=X^c_i\sum_qE^c_{i,q}z^q$. Both sides are polynomials in $z$ of
degree less than $G=H/w$; unless they are equal as polynomials, a random $z$ satisfies the equation
with probability at most $(G-1)/|\K|$. Equality as polynomials means $X^c_iE^c_{i,q}=X^S_{ij^*}[q=g^S_{ij^*}]$
for every $q$, hence $\bar c_i=S_{ij^*}$ and each product $X^c_iE^c_{i,q}$ is bounded by $\Xmax$. Every
consumer of $c$ reads it only through these products (the limbs of \cref{lem:elem-int}), thus it
sees the value $S_{ij^*}$. By (iii), $\RN(\bar c_i-S_{ij})\ge0$ for every $j\le i$. Since $\bar c_i$ and
$S_{ij}$ are multiples of $2^{e_S}$, a negative difference has magnitude at least $2^{e_S}\ge2^{e_\Delta}$
and rounds to a negative value, hence $\bar c_i\ge S_{ij}$ for all $j\le i$.
\end{proof}

\subsection{Proof of \cref{thm:block}}\label{sup:prf:thm:block}
\begin{proof}
Completeness is that of each component: the honest prover computes every tensor by its definition,
hence every product identity, gadget, lookup and selector check holds, and the openings are
consistent because each commitment is opened only at points at which its own polynomial was
evaluated. For soundness, suppose no component check fails beyond its error. The certification of
the input and the parameters (one-hot indicators, limbs in $\mathcal L$) shows that they are valid,
though not necessarily canonical, encodings of elements of their formats, which is condition (i).
Proceed along \cref{tab:steps}. When a step is reached, each of its operands is, by induction, a
valid encoding of its defined value (the input or a parameter), a canonical encoding (a tensor
produced earlier), or the auxiliary factorization of a maximum. By \cref{lem:view} the step's checks
are checks about exactly these committed polynomials. Matrix products and element-wise operations
accept valid encodings and factorizations (\cref{lem:bucket,lem:elem-int}), and every unary lookup
reads a tensor produced by a rounding gadget, hence a canonical one. The lemma cited for the step
then shows that its output is the canonical encoding of its defined value, which is (ii) and (iv).
For a maximum, \cref{lem:max} shows that its products equal those of the selected score and that
the selected score is the row maximum, which is (iii); the difference gadget that consumes it reads
only these products. Hence all tensors, and in particular $x_2$, are the canonical block of $x$
and the parameters in the sense of the definition.
\end{proof}

\subsection{Proof of \cref{lem:zkbind}}\label{sup:prf:lem:zkbind}
\begin{proof}
The masking rows are rows of the committed matrix and appear in the proximity combination $u_0$ with
their own powers of $\vartheta$, hence the proximity test and the column checks are those of the
underlying scheme and bind every row, masking rows included, to a codeword. Let $\bar M_k^{\K}$ be the
decoded masking combination and $\bar v_{k,p}$ the true evaluations of the decoded tables. The
values $v$ and $\mu$ are fixed before $\lambda$, and \eqref{eq:zkopen} with the decoded rows reads
$\sum_p\lambda^p(\bar v_{k,p}-v_{k,p})+\lambda^{n_p}(\langle\eq(b_k,\cdot),\bar M_k^{\K}\rangle-\mu_k)=0$.
If some $v_{k,p}$ is wrong, this is a non-zero polynomial of degree at most $n_p$ in $\lambda$, which
vanishes at a uniform $\lambda\in\K\setminus\Fp$ with probability at most $n_p/(|\K|-p)$. The mask term
has its own power of $\lambda$, which prevents an announced mask value from absorbing an error in
$v_{k,0}$.
\end{proof}

\subsection{Proof of \cref{lem:zkhide}}\label{sup:prf:lem:zkhide}
\begin{proof}
Since $\tilde{\hat f}_p(z_k)=(1-y_k)\tilde f_p(z_k^\circ)+y_k\tilde\pi_p(z_k^\circ)$ with $\pi_p$ the
pad of table $p$, uniform and independent across $p$, and $y_k\ne0$, \eqref{eq:rank} makes
$(v_{k,p})_k$ uniform for each $p$, independently of $f$. The masking rows are independent of the
pads. The row $M_k^{\K}$ is uniform in $\K^c$ and $\eq(b_k,\cdot)\ne0$, hence $\mu_k$ is uniform in $\K$;
conditioned on $\mu_k$, $M_k^{\K}$ is uniform on the hyperplane $\langle\eq(b_k,\cdot),M\rangle=\mu_k$,
hence $u_k$ is uniform on the hyperplane \eqref{eq:zkopen}, and the value $\mu_k$ reveals nothing
beyond that relation. The rows $M_0,M_0'$ occur only in $u_0$, with coefficients
$\vartheta^{i},\vartheta^{i+1}$, whose $\Fp$-span is $\K$ because $\vartheta\notin\Fp$; hence $u_0$ is
uniform. The extension parts of all $u$'s are masked in the same way by the extension symbols of
the masking rows. Finally, conditioned on the revealed combinations, the extension symbols of every
row are uniform on an affine subspace whose direction contains, for each row, all extension vectors
of that row orthogonal to the revealed combinations; by the rank statement for the encoder, the
symbols at the at most $Q<e$ distinct queried positions are then uniform subject to exactly the
linear equations that the verifier checks.
\end{proof}

\subsection{Proof of \cref{thm:zk}}\label{sup:prf:thm:zk}
\begin{proof}
(a) and (b). Completeness and soundness follow as for \cref{thm:block}. The new ingredients are
sound by \cref{lem:zkbind,lem:zksc,lem:pool}, the zero-sum forms are equivalent on the data half to
the checks they replace, the gather and output zero sums are zero-checks with the error of
\cref{lem:zerocheck} plus $G/|\K|$ for the combination with $\kappa$, and the phases compose because a
carried commitment is the same polynomial in every phase that reads it, hence claims made in
different phases about it are discharged by its single opening. By induction over the phases in the
order of the computation, every tensor satisfies the conditions of \cref{def:block} for its step,
the embedding rows are the rows of the committed vocabulary at the public tokens, and the last row
of the logits equals $y$.

(c). The simulator $\mathcal{S}(\mathcal{P},\tau,y)$ proceeds phase by phase. A Merkle root is the
oracle answer on the root node; the simulator outputs a uniform string and later programs the
oracle along the authentication paths of the opened columns, consistently with simulated column
contents. This fails only if the adversary has queried the oracle on an input that the simulator
later programs (a leaf message or the pair of child hashes of an internal node), which is the event
bounded by $\mathsf E(\varepsilon_{\mathrm{RO}})$. The simulator samples the verifier's coins and, under
$\mathsf H$, samples the vector of all revealed evaluations and product factors uniformly. For
every masked sumcheck it then knows the claim (zero, or a previously sampled factor) and the value
$F(q)$, which the verifier computes from revealed values, and it simulates the messages by
\cref{lem:zksc}, using $\rho\ne0$. It simulates every pool opening as in \cref{lem:pool} and every
commitment opening by sampling mask values, row combinations and columns as in \cref{lem:zkhide}.
Public values (the compressed tables, the public output, the public constants) are computed from
the statement. Each component of the real view has, conditioned on the previous ones, the
distribution sampled by $\mathcal S$ unless $\mathsf H$ fails, a rank condition \eqref{eq:rank} fails
(at most $2\nu K/p$ per commitment when the conditional statement after \cref{lem:zkhide} applies), a pad coordinate $y_k\in\{0,1\}$ (at most $1/p$ per point, as
$y_k$ is uniform in $\K$ and the bound is taken over its base-field components), or the oracle
simulation fails; the union bound gives $\varepsilon_{\mathrm{zk}}$.
\end{proof}

%% file: supp_zk.tex
\subsection{The Rank Condition}\label{sup:rank}
The rank condition \eqref{eq:rank} is a genuine hypothesis and does not follow from the points being
distinct. It fails, for example, if two points share their reduced part $z^\circ$ but have different
pad coordinates $y_1\ne y_2$ (then $(y_2v_1-y_1v_2)/(y_2-y_1)$ is the data evaluation), if a reduced
point is entirely Boolean (its pad evaluation lies in $\Fp$), or if two reduced points are conjugate
under the Frobenius map. Claims that coincide (such as the two operands of $\sum_jz_j^2$) are treated as one.

Views that fix coordinates restrict the randomness that a claim can use, and no bound that ignores
them is valid. If two points fix the same reduced coordinates to Boolean values and leave $d$
coordinates free, their pad evaluations depend only on the $2^d$ pad entries of that Boolean slice;
for instance, with $\nu=2$ and points $(0,a)$, $(0,b)$, $a,b\in\K$, the evaluations
$(1-a)\pi_{00}+a\pi_{01}$ and $(1-b)\pi_{00}+b\pi_{01}$ involve two base-field unknowns, hence $\Phi$
has rank at most $2<2\cdot2$ and \eqref{eq:rank} fails for all $a,b$. A necessary condition is
therefore that $m$ points restricted to a common slice with $d$ free coordinates satisfy
$2m\le2^d$. What does hold is the following conditional statement: if, for the actual views and the
actual sharing of challenges, some assignment of the free coordinates makes $\Phi$ onto, then the
determinant of a $2K\times2K$ minor of $\Phi$ is a non-zero polynomial of degree at most $2K\nu$ in the
base-field components of the free coordinates, and \eqref{eq:rank} fails with probability at most
$2K\nu/p$ over them (the bound is in $1/p$, not $1/|\K|$). We have not exhibited such an assignment for
every commitment of the complete schedule; \eqref{eq:rank} is therefore part of the hypotheses of the
zero-knowledge statement.

\subsection{Proof of \cref{lem:pool}}\label{sup:prf:lem:pool}
\begin{proof}
If a claimed value is wrong, $T\ne\langle W,U\rangle$ except with probability $S_\mathrm{c}/|\K|$
over $\lambda$. Since $\tau$ is fixed before $\rho'$ and the column checks bind $u$ to $U+\rho' d$,
the check reads $\langle W,U\rangle-T+\rho'(\langle W,d\rangle-\tau)=0$, which fails for all but one
$\rho'$. For hiding, $d$ is uniform in $\K^{8S}$ and independent of $U$, $W\ne0$ and $\rho'\ne0$; hence
$\tau$ is uniform and, given $\tau$, $u$ is uniform on the hyperplane $\langle W,u\rangle=T+\rho'\tau$.
The columns are handled as in the proof of \cref{lem:zkhide}: each of the four rows carries
$e>Q$ extension symbols, and $u$ fixes one $\K$-combination of them.
\end{proof}
\subsection{On the Masking Hypothesis}\label{sup:hyp}
A sufficient condition is a \emph{fresh-pad order}: an ordering $V_1,\dots,V_{N_{\mathrm v}}$ of the
revealed values such that, for every $i$, conditioned on the pads that occur in $V_1,\dots,V_{i-1}$,
the value $V_i$ is an affine function, onto $\K$, of pad entries that occur in no earlier value. Then
each $V_i$ is uniform given its predecessors, and $V$ is uniform. Indicator evaluations would be placed
before the factors that multiply indicator pads by limb pads; the ordering must respect the
fixed-coordinate restrictions discussed after \cref{lem:zkhide}. For a single commitment and
evaluations only, $\mathsf H$ reduces to \eqref{eq:rank}. In general it couples all claims about all
commitments of the transcript, in all phases, and the factors that reuse the same limb and indicator
pads. We have not constructed a fresh-pad order for the complete schedule, and we do not bound
$\varepsilon_{\mathsf H}$ in this paper. We state the zero-knowledge property under $\mathsf H$
(\cref{thm:zk}); proving it, or replacing it by independent masks per revealed value, is the main
open step of the security analysis.

\subsection{Leakage of Formats and Binding to a Published Model}\label{sup:zkmore}
The statement reveals the prompt, the logits of the last position, the
shapes, and the format of every tensor, hence an upper bound on the magnitude of its entries at
the granularity of $w=12$ bits (Section~\ref{sup:ext}). In our experiments the formats were chosen from
this very input; a deployment must fix them from calibration data, independent of the input, or
they leak coarse information about it.

The parameters are committed afresh in each proof.
To bind many proofs to one published commitment of the parameters, that commitment must support
openings in all of them: with $Q$ queries per proof and $N$ proofs this needs $e\ge NQ+1$ extension
symbols per row and $2N(K+1)$ masking rows, and the pads must stay unrevealed across proofs, which
holds by \cref{lem:zkhide} only if the union of all opening points of all proofs satisfies the rank
condition \eqref{eq:rank}; the number of proofs a commitment supports is therefore bounded by its
pad dimension. We have not implemented this variant.

%% file: supp_ext.tex
\section{Extensions and Discussion}\label{sup:ext}

\Cref{sec:elem} covers every operation of a GPT-2 block. Other architectures need the same
ingredients: RMSNorm is LayerNorm without the mean, rotary embeddings are products with public
tables, and gated activations are a lookup and a product. The semantic experiments of
\cref{sec:eval} (perplexity, kernels) evaluate element-wise layers in binary64 with one rounding,
which differs from \cref{def:ln,def:sm} in where intermediate roundings occur; the block of
\cref{sec:blockrun} is evaluated in the canonical form.

\emph{Faithful rounding and native kernels.} Replacing $a<4P$ in $\ta$ by $a<4P(1+2\eta)$ and shifting
$V$ by $\eta U$ proves $|C_{ij}-v_{ij}|\le(\tfrac12+\eta)u$, a relative faithful rounding; it accepts
native kernels only when their error is below $\eta u$, which fails under cancellation, and
Section~\ref{sup:slack} shows that a kernel-agnostic binary32 bound lets a prover steer the generated
token. Proving a fixed native kernel bit for bit by tracing only undetermined entries still costs $k$
trace steps per such entry, and with the bound of \cref{prop:higham} $1.5\%$--$99.7\%$ of the entries of
GPT-2 products are undetermined for binary32 (\cref{tab:e1}), although fewer than $0.03\%$ differ; a
certifiable order-specific bound is the main open problem on the native side.

Apart from the $T_D$ limbs and one inverse per entry, every
committed value is a bit or has at most $w+8$ bits. Commitments whose cost grows with the norm of
the committed vector, such as Ajtai-type commitments in lattice-based proof
systems, are therefore a natural
target, as are binary-field commitments~\cite{binius23}. We leave these instantiations to
future work.

Windowed formats reveal, per tensor, an upper bound on the
magnitude of its entries, at a granularity of $w$ bits. Unrestricted \bfs{} ($H=254$) would reveal
nothing, at a cost linear in the number of buckets, but it is not an instance of \cref{thm:main} as
stated (\cref{rem:general}); it requires the alignment and final-bucket analysis that we omit, and we
do not claim it.

\begin{remark}[General alignment]\label{rem:general}
The development above assumes \eqref{eq:align} and $w\mid H_\bullet$. A partial generalization is as
follows. If $w\nmid(e_C-e_A-e_B)$, let $s\in[w]$ with $w\mid(e_C-e_A-e_B+s)$ and replace $N$ by
$2^sN$, that is, multiply the limbs $D^{(t)}$ in (C7) by $2^s$. This is valid when, in addition,
$(e_C-e_A-e_B+s)/w\ge1$ and the bias offset satisfies $e_b-e_A-e_B+s\ge0$ and
$w\mid(e_b-e_A-e_B+s)$; the proofs then go through with $\Emax$ multiplied by $2^s$. When these
conditions fail, one chooses a common integer unit below all offsets, decomposes each offset into a
bucket shift and a within-bucket bit shift, and re-derives the limb and carry bounds; we do not
carry this out. If $w\nmid H$, one may take $G=\lceil H/w\rceil$ buckets and restrict the last bucket
to valid exponents by an additional constraint; for full \bfs{} ($H=254$, $e_C-e_A-e_B=133$) no
$w\ge2$ divides both, and full \bfs{} is therefore not an instance of \cref{thm:main} as stated.
Our experiments use windowed formats, for which \eqref{eq:align} and $w\mid H$ hold.
\end{remark}

\begin{remark}[Other precisions]\label{rem:precision}
Nothing depends on the significand having 8 bits except the size of $\tenc$, which is
$2^{\pi}w+2^{\pi}-1$ rows for a $\pi$-bit significand (for $\pi=8$, the $256w+255$ rows of (L1)), and
the bound $\Xmax<2^{\pi+w-1}$. Formats with
$\pi\le11$ (binary16, the FP8 formats) need only a different table. For binary32 the significand is
split into limbs, at the cost of a few more cells per output.
\end{remark}

%% file: supp_eval.tex
\subsection{Exponent Ranges}\label{sup:exp}

The cost of the protocol is linear in the number of exponent buckets, thus the spread of exponents
matters. Across all operands and results of all matrix products (same data as \cref{tab:e1}), the
span between the smallest and largest binary exponent of non-zero entries, plus the eight
significand bits, is between 31 and 45 bits, with two exceptions: GELU outputs span
65 bits, with a fraction $7\cdot10^{-5}$ of the entries more than 48 bits below the maximum, and
attention probabilities span 141 bits, with $1.8\%$ of the entries below a 48-bit window. These
tails are values such as $\mathrm{GELU}(-10)\approx-8\cdot10^{-23}$ and softmax outputs of strongly
disfavoured positions.

Exact \bfs{} therefore needs $G=4$ buckets of 12 bits for most operands, 6 for GELU outputs and 12
for attention probabilities. Windowed formats with $G=4$ everywhere replace the tails by gradual
underflow. On the evaluation data of \cref{sec:model} this changes $0.70\%$ of the entries of
element-wise results ($3{,}234{,}149$ of $459{,}276{,}288$), all of them in the tails; exactly one
of $734{,}334{,}976$ matrix-product outputs falls in the underflow zone of its window; no value
exceeds its window; and the final logits are bit-identical to those of unwindowed \bfs{} at all
$4{,}096$ positions.

\subsection{What an Error-Bounded Relation Would Give Away}\label{sup:slack}

An error-bounded relation lets every output be any $\RN(x)$ with $|x-v_{ij}|\le\beta_{ij}$ (\cref{prop:higham}).
Our adversary back-propagates the margin between the runner-up and the greedy logit to every product
output (straight-through estimator for roundings), claims the upper or lower admissible value by the
sign of the gradient, re-runs the forward pass with these claims, and repeats three times; element-wise
layers are honest and final logits are computed in binary64. The attack is first-order, hence a lower
bound on what a dishonest prover can do.

\begin{table}[!htb]
\centering\footnotesize\setlength{\tabcolsep}{2.5pt}
\caption{Adversarial slack of the error-bounded relation on GPT-2: 48 targets (the last position of
48 sequences of 32 tokens). The median margin between the two largest logits of the honest
execution is $0.72$. ``Random choices'' picks the upper or lower admissible value at random and
measures the background sensitivity of \bfs{} inference.}\label{tab:slack}
\begin{tabular}{lrrrrr}
\toprule
roundoff & with & token & \multicolumn{2}{c}{margin gain} & random\\
\cmidrule(lr){4-5}
$\mathsf{u}$ & choice & flipped & median & max & flips\\
\midrule
$2^{-24}$ (binary32) & $40.8\%$ & $47/48$ & $3.50$ & $9.13$ & $1/48$\\
$2^{-28}$            & $8.3\%$  & $9/48$  & $0.24$ & $1.27$ & $0/48$\\
$2^{-32}$            & $0.96\%$ & $1/48$  & $0.03$ & $0.43$ & $0/48$\\
$2^{-53}$ (binary64) & $0.007\%$& $0/48$  & $0.01$ & $0.34$ & $1/48$\\
\bottomrule
\end{tabular}
\end{table}

With the binary32 bound the adversary flips the greedy token at 47 of 48 targets, including 8 of the
9 with honest margin above two logits, while random choices in the same slack flip one: the damage
comes from coordination of one-ulp deviations. Such a relation becomes meaningful only with bounds
about eight bits tighter than the worst case; the canonical relation has no slack.

\subsection{The Protocol on Real Operands}\label{sup:piop}

\begin{table}[!htb]
\centering\footnotesize\setlength{\tabcolsep}{2.5pt}
\caption{Reference prover and verifier on slices of real GPT-2 operands from layer 5 under
\textsf{CR-w} ($G_A=G_B=G_C=4$, $T_D=7$, $w=12$, $n_c=4$; $k$ is padded to a power of two). Cells and lookups are per output; MACs are multiply--adds.
``cells/out'' is $n_{\mathrm{cell}}$ of \eqref{eq:cells}. The Python prover took between $0.4$ and
$15$ seconds and the verifier between $0.1$ and $1.6$ seconds; these times characterize an
unoptimized interpreter, not the protocol.}\label{tab:piop}
\begin{tabular}{lrrrrrrrrr}
\toprule
site & $m$ & $k$ & $n$ & $d$ & $T_c$ & cells & lookups & MACs & verdict\\
\midrule
\textsf{qk}    & 32 & 64   & 32  & 3 & 7 & 52 & 32 & $6.6\cdot10^4$ & accept\\
\textsf{av}    & 32 & 32   & 64  & 3 & 7 & 52 & 32 & $6.6\cdot10^4$ & accept\\
\textsf{fc}    & 16 & 768  & 64  & 4 & 8 & 57 & 37 & $7.9\cdot10^5$ & accept\\
\textsf{fc}    & 32 & 768  & 256 & 4 & 8 & 57 & 37 & $6.3\cdot10^6$ & accept\\
\textsf{aproj} & 32 & 768  & 256 & 5 & 9 & 62 & 42 & $6.3\cdot10^6$ & accept\\
\textsf{mproj} & 16 & 3072 & 128 & 5 & 9 & 62 & 42 & $6.3\cdot10^6$ & accept\\
\textsf{qkv}   & 32 & 768  & 512 & 5 & 9 & 62 & 42 & $1.3\cdot10^7$ & accept\\
\bottomrule
\end{tabular}
\end{table}

\Cref{tab:piop} lists seven instances built from operands
captured during a \textsf{CR-w} forward pass, with inner dimension up to $3072$. The verifier
accepts all of them, and also 13 hand-built and 12 random honest instances.

As a larger test we proved every matrix product of one
transformer block: all 28 products of layer~5 of GPT-2 at full width for a 16-token input (the
input projection, twelve query--key products, twelve attention-weighted value products, and the
three remaining projections). Together they have $125{,}952$ output entries and $1.14\cdot10^{8}$
multiply--adds. The verifier accepts all 28. The proofs commit to $7.4\cdot10^{6}$ cells in the
output tables, against $7.3\cdot10^{9}$ constraints at 64 per multiply--add. The Python prover
needed $278$ seconds in total and the verifier $32$ seconds. The products are proved separately on
the same data; the commitments are not chained across products here. \Cref{sec:blockrun} proves a
complete block, element-wise layers included, with its tensors chained.

Every false statement and malformed witness of \cref{tab:tests} is rejected; these tests exercise the
case analysis of \cref{lem:round,lem:gadget} and are evidence that the implementation matches the
proofs, not a substitute for them.

\subsection{Cost and Comparison with the State of the Art}\label{sup:cost}

\begin{table}[!htb]
\centering\scriptsize\setlength{\tabcolsep}{2pt}
\caption{Cost model for all matrix products of one GPT-2 forward step (context 128): committed cells against constraints of per-operation gadgets at 64 and 90 per multiply--add.}\label{tab:cost}
\begin{tabular}{lrrrrrr}
\toprule
 & & outputs & MACs & this work & \multicolumn{2}{c}{gadgets}\\
\cmidrule(lr){6-7}
site & $k$ & /token & /token & (cells) & 64 & 90\\
\midrule
\textsf{qkv}    & 768  & 27{,}648 & $2.1\cdot10^7$ & $1.6\cdot10^6$ & $1.4\cdot10^9$ & $1.9\cdot10^9$\\
\textsf{qk}     & 64   & 18{,}432 & $1.2\cdot10^6$ & $1.1\cdot10^6$ & $7.5\cdot10^7$ & $1.1\cdot10^8$\\
\textsf{av}     & 128  &  9{,}216 & $1.2\cdot10^6$ & $5.3\cdot10^5$ & $7.5\cdot10^7$ & $1.1\cdot10^8$\\
\textsf{aproj}  & 768  &  9{,}216 & $7.1\cdot10^6$ & $5.7\cdot10^5$ & $4.5\cdot10^8$ & $6.4\cdot10^8$\\
\textsf{fc}     & 768  & 36{,}864 & $2.8\cdot10^7$ & $2.1\cdot10^6$ & $1.8\cdot10^9$ & $2.5\cdot10^9$\\
\textsf{mproj}  & 3072 &  9{,}216 & $2.8\cdot10^7$ & $5.7\cdot10^5$ & $1.8\cdot10^9$ & $2.5\cdot10^9$\\
\textsf{logits} & 768  & 50{,}257 & $3.9\cdot10^7$ & $2.9\cdot10^6$ & $2.5\cdot10^9$ & $3.5\cdot10^9$\\
\midrule
total           &      & 160{,}849 & $1.26\cdot10^8$ & $9.4\cdot10^6$ & $8.1\cdot10^9$ & $1.1\cdot10^{10}$\\
\midrule
ratio to this work & & & & $1$ & $861$ & $1211$\\
\bottomrule
\end{tabular}
\end{table}

By \eqref{eq:cells}, the protocol commits to $n_{\mathrm{cell}}$ field
elements per output entry, independent of the inner dimension $k$. For the calibrated formats of
GPT-2, $n_{\mathrm{cell}}$ is between 52 and 62, of which 7 are limbs of up to 51 bits, one is a
full field element, and the rest are bits or values of at most 21 bits; there are 32--42 lookups
per entry. A per-operation approach pays for every multiply--add: $64$ constraints per
single-precision multiplication in the most optimized IEEE-compliant circuits~\cite{ernstberger24},
and about $90$ constraints per \bfs{}/binary32 multiply--accumulate in the gadget
of~\cite{peigne26}. \Cref{tab:cost} evaluate both for one forward step of GPT-2: the
protocol commits to $9.4\cdot10^6$ cells per token against $8.1\cdot10^9$ and $1.1\cdot10^{10}$
constraints, a ratio of $861$ and $1211$. The ratio is $64k/n_{\mathrm{cell}}$ per product and
grows linearly with the inner dimension, thus it is larger for wider models.

Cells and constraints are only proxies for prover time, for different relations; integer zkML pays a
handful of cells per output for rescaling, our $52$--$62$ are the price of exact floating point.
Published proving times are on different hardware and for different relations: DeepProve~\cite{deepprove26} reports 174 tokens per minute for GPT-2 on 16--24 cores, zkLLM~\cite{zkllm24} 803~s for Llama-2 13B on one A100 GPU, SASH~\cite{sash26} 1--1.5~h for Llama-3.1 8B on one H100, FlashProve~\cite{flashprove26} 110 tokens per second for a 400B mixture of experts on one H100, and ZIP~\cite{zip25} 34.5~h on 48 cores for the linear layers of an 11M-parameter model. Against ZIP this is more than three orders of magnitude fewer core-seconds on a model eleven times
larger, the gap that the removal of the cubic term predicts, although the relations differ. Against
DeepProve (about 7 to 11 tokens per minute per core) our marginal cost of $0.61$ seconds per token
and block corresponds to roughly 4 tokens per minute per core for the matrix products of twelve
blocks, an extrapolated order of magnitude: exact floating point costs a small constant factor over
an integer model.

\subsection{Order Dependence of Native Kernels}\label{sup:order}
\begin{table}[!htb]
\centering\footnotesize\setlength{\tabcolsep}{3pt}
\caption{Order dependence of binary32 accumulation: fraction of output entries, in units of
$10^{-4}$, on which two kernels differ after rounding to \bfs{}. L: strict left-to-right; R: strict
right-to-left; BLAS: library order; CR: correctly rounded. One sequence of 48 tokens; between
$36{,}864$ and $147{,}456$ entries per row, hence individual cells rest on between 5 and 23 differing
entries.}\label{tab:e2}
\begin{tabular}{lrrrrrr}
\toprule
site & $k$ & L$\ne$R & L$\ne$BLAS & L$\ne$CR & R$\ne$CR & BLAS$\ne$CR\\
\midrule
layer 0, \textsf{fc}     & 768  & $0.8$ & $0.6$ & $0.9$ & $0.5$ & $0.7$\\
layer 0, \textsf{mproj}  & 3072 & $4.1$ & $3.0$ & $3.5$ & $3.5$ & $1.6$\\
layer 5, \textsf{qkv}    & 768  & $2.1$ & $0.8$ & $1.5$ & $1.5$ & $1.4$\\
layer 5, \textsf{aproj}  & 768  & $3.8$ & $4.9$ & $3.3$ & $4.6$ & $4.6$\\
layer 5, \textsf{fc}     & 768  & $1.5$ & $0.9$ & $1.1$ & $1.4$ & $0.4$\\
layer 5, \textsf{mproj}  & 3072 & $5.2$ & $4.9$ & $4.6$ & $4.9$ & $3.3$\\
layer 11, \textsf{fc}    & 768  & $0.9$ & $0.3$ & $0.6$ & $0.9$ & $0.7$\\
layer 11, \textsf{mproj} & 3072 & $5.4$ & $3.8$ & $3.8$ & $3.5$ & $2.4$\\
\bottomrule
\end{tabular}
\end{table}

\Cref{tab:e2} compares three native summation orders with one another and with the canonical
result: two native orders disagree with each other as often as either disagrees with the correctly
rounded result, so the canonical semantics is one more native kernel, defined without reference to an
order, while a relation pinning a native kernel bit for bit must fix order, blocking and device.

\subsection{Design Flaws Found During Development}\label{sup:attacks}
Four flaws of our first design are easy to make: (a) masking revealed row combinations with pad rows
alone reveals a random combination of data evaluations, hence dedicated masking rows; (b) a masking row
with coefficient $\lambda^0$ can absorb an error in the first polynomial (soundness), hence its own power;
(c) one $\Fp$ masking row covers half of a $\K$-combination; (d) a code length rounded from $2(c+e)$ gives
rate up to $1/2$ and only $2^{-66}$ proximity error. An audit found three more: separately announced
$\Fp$ mask values reveal a data functional (one $\K$-valued masking row with a single value is used);
$\lambda,\vartheta$ must lie outside $\Fp$; and $\rho,\rho'$ must be non-zero. All measurements use the
corrected prover.

%% file: appendix.tex
\section{Parameters of the Reference Implementation}\label{app:params}

\begin{table}[t]
\centering\scriptsize\setlength{\tabcolsep}{2pt}
\caption{Parameters. For $k\le8192$, $\Emax<2^{53}$, hence
$\Emax+(\beta+1)\beta^{n_c}/2<2^{60}<p$ and $\Emax/(\beta-1)<2^{41}<\beta^{n_c}/2=2^{47}$, which is
\eqref{eq:pc}; and $2k\Xmax^2<2^{53}<p$.}
\begin{tabular}{@{}lp{3.6cm}l@{}}
\toprule
symbol & meaning & value in the experiments\\
\midrule
$p$ & field modulus & $2^{64}-2^{32}+1$\\
$w$, $\beta$ & bucket width, base & $12$, $4096$\\
$H_\bullet$, $G_\bullet$ & exponent range and buckets per format & $48$, $4$\\
$\Xmax$ & largest limb magnitude, $255\cdot2^{w-1}$ & $522{,}240$\\
$T_D$ & product limbs, $G_A+G_B-1$ & $7$\\
$d$ & $(e_C-e_A-e_B)/w$ & $3$ to $5$\\
$T_c$ & chain length, $\max(T_D,d+G_C,d_b+G_b)$ & $7$ to $9$\\
$T_b$ & residual digits, $d+G_C-2$ & $5$ to $7$\\
$n_c$ & carry digits & $4$\\
$|\tenc|$, $|\ta|$, $|\rbeta|$, $|\rquart|$ & table sizes & $3327$, $16380$, $4096$, $1024$\\
\bottomrule
\end{tabular}
\end{table}

The witness generator computes the limbs $D^{(t)}$ with binary64 matrix products of the integer
limb matrices, which is exact because $k\Xmax^2<2^{53}$ for $k\le8192$.

\section{Tests and Reproducibility}\label{app:tests}\label{app:repro}

\begin{table}[!h]
\centering\scriptsize\setlength{\tabcolsep}{2pt}
\caption{Tests and attacks; all verdicts are as expected (acc./rej.).}\label{tab:tests}
\begin{tabular}{@{}p{6.3cm}rl@{}}
\toprule
test & count & verdict\\
\midrule
reference: honest (random, \cref{tab:piop}, layer 5, ties, binade, cancellation, underflow) & 60 & acc.\\
reference: wrong output/neighbour, sign, zero; altered limb, carry, $a$, $\tp$, $E_0$, $P$, $b_0$ & 18 & rej.\\
reference, randomized: outputs off by 1--3 units; re-balanced limbs & 239 & rej.\\
native prover: all GPT-2 products; layer 5 at 4/16/64 tokens / 3 tamperings & 427/3 & acc./rej.\\
block: honest / tensor changed / wrong layer, consistent later / proof changed & 1/26/19/10 & acc./rej.\\
complete model: unit tests; honest (1 and 12 blocks) & 4/2 & acc.\\
complete model: tensor changed / consistent wrong run / proof changed (1 / 12 blocks) & 25/8/161/42 & rej.\\
\bottomrule
\end{tabular}
\end{table}

The code contains the reference semantics and protocol (\texttt{crfloat.py}, \texttt{gpt2.py},
\texttt{piop.py}), the experiment and attack scripts, and the native, block and full-model provers
(\texttt{prover/}, \texttt{block.rs}, \texttt{prover3/}, \texttt{zkmodel}, \texttt{model\_ref.py}). Every
number in \cref{sec:eval} is read from the JSON files these programs write. The \textsf{native} variants
reflect a CPU binary32 kernel; rates should be re-measured on the accelerator of interest.